\documentclass{article} 
\PassOptionsToPackage{nonamebreak}{natbib}
\usepackage{iclr2027_conference,times}

\usepackage{amsmath,amsfonts,bm}

\def\eqref#1{equation~\ref{#1}}

\def\1{\bm{1}}

\DeclareMathAlphabet{\mathsfit}{\encodingdefault}{\sfdefault}{m}{sl}
\SetMathAlphabet{\mathsfit}{bold}{\encodingdefault}{\sfdefault}{bx}{n}

\usepackage{hyperref}
\usepackage{url}
\usepackage{graphicx}
\usepackage{cleveref}
\usepackage{amssymb}
\usepackage{amsmath}
\usepackage{booktabs}
\usepackage[dvipsnames]{xcolor}
\usepackage{caption}
\usepackage{placeins}
\usepackage{float}
\usepackage{needspace}

\ifdefined\GlassPaperCommandsLoaded
\else
\def\GlassPaperCommandsLoaded{}

\providecommand{\glassrows}[1]{\csname @@input\endcsname #1 }
\providecommand{\glassfit}[1]{%
  \begingroup
  \sbox{0}{#1}%
  \ifdim\wd0>\linewidth
    \resizebox{\linewidth}{!}{\usebox{0}}%
  \else
    \usebox{0}%
  \fi
  \endgroup
}

\newcommand{\QmofSamplingSteps}{32}
\newcommand{\MpTableCheckpoint}{50k}
\newcommand{\MpTableNote}{}

\newcommand{\MpTrainCellRmsd}{0.0339}

\newcommand{\MpValCellRmsd}{0.3001}

\newcommand{\QmofTrainCellRmsd}{0.0190}
\newcommand{\QmofValCellRmsd}{1.2473}

\newcommand{\QmofValidity}{74.8}
\newcommand{\QmofTrainingValidity}{80.3}
\newcommand{\QmofRelaxedValidity}{78.72}

\newcommand{\QmofLargeValidity}{67.8}
\newcommand{\QmofTrainingLargeValidity}{78.5}
\newcommand{\MofasaLargeValidity}{38.3}
\newcommand{\ZatomLargeValidity}{0.8}

\newcommand{\QmofVnu}{8.76}

\newcommand{\QmofStrictVnu}{2.70}

\newcommand{\QmofEarlyValidUnmatchedYield}{2.20}
\newcommand{\QmofTrainingMatchRate}{98.3}
\newcommand{\QmofValidUnmatchedYield}{1.28}
\newcommand{\QmofUnmatchedNewFormula}{99.7}

\newcommand{\MofasaAnySplitMatchRate}{0.15}

\newcommand{\LematBulkTrainCellRmsd}{0.043}
\newcommand{\LematBulkValCellRmsd}{0.046}
\newcommand{\LematBulkTrainStructures}{5,101,846}

\fi

\title{GLASS: Global Latent Aggregation with Slot-based Set Decoding for Scalable All-Atom Crystal Generation}

\author{%
Hendrik Kra\ss\thanks{\texttt{hendrik.krass@ki.uni-stuttgart.de}} \\
Institute for Artificial Intelligence \\
University of Stuttgart
\And
Seyed Mohamad Moosavi\thanks{Joint last author} \\
Department of Chemical Engineering \\
\& Applied Chemistry \\
University of Toronto \\
Vector Institute
\AND
Mathias Niepert\footnotemark[2] \\
Institute for Artificial Intelligence \\
University of Stuttgart \\
NEC Labs Europe
}

\iclrfinalcopy 
\begin{document}

\maketitle

\begin{abstract}
Generative models for crystals enable the discovery of novel structures, but scaling all-atom generation to larger systems such as metal--organic frameworks remains challenging. We connect this difficulty to the correspondence problem of particle-space generation. Even on a single fixed target set, index-free permutation-equivariant particle flows require substantially more training for reliable generation as set size and density increase, under both independent and optimal-transport couplings. To resolve this challenge, we introduce GLASS---Global Latent Aggregation with Slot-based Set Decoding, which encodes structures in a permutation-invariant global latent space and learns their distribution via flow matching. A learned-slot decoder constructs all atoms in parallel, removing atom-wise correspondence from generative transport. On MP20, GLASS is competitive with particle-space models, and flow training can reach the validity of the training data at every structure size. On a QMOF subset, GLASS generates MOFs with up to 150 atoms per unit cell without conditioning on building blocks, topology, or composition, and approaches the structural validity of the training data. On both datasets, flow training exposes a validity--novelty tradeoff, and MOF novelty remains limited by autoencoder generalization on the available data. These results show that separating correspondence assignment from generative transport provides a simple route toward high-validity generation of larger atomistic systems. 
\end{abstract}

\section{Introduction}

Generative models for crystals are increasingly capable of producing valid and novel structures \citep{xieCrystalDiffusionVariational2022, millerFlowMMGeneratingMaterials2024, betalaLeMatGenBenchUnifiedEvaluation2025}. This is particularly promising for materials discovery, where useful structures occupy only a tiny fraction of a vast chemical design space \citep{sanchez-lengelingInverseMolecularDesign2018}, and novel materials are required to address challenges such as carbon capture \citep{ozkanStatusProspectsMaterials2022}, water purification \citep{werberMaterialsNextgenerationDesalination2016}, catalysis \citep{guoRationalDesignEarthAbundant2024}, and energy storage \citep{taborAcceleratingDiscoveryMaterials2018}. However, these models are frequently evaluated on datasets with limited structure size, such as MP20 \citep{xieCrystalDiffusionVariational2022, jainCommentaryMaterialsProject2013} and MPTS52 \citep{bairdMatbenchgenmetricsPythonLibrary2024}, which contain structures with up to 20 and 52 atoms, respectively.

Metal--organic frameworks (MOFs), with applications such as carbon capture, are an important target beyond small-crystal benchmarks \citep{duanRiseGenerativeAI2025}. Many approaches simplify MOF generation by decomposing structures into reusable building blocks \citep{fuMOFDiffCoarsegrainedDiffusion2024, jiaoMOFBFNMetalOrganicFrameworks2025, duanBuildingBlockAwareGenerative2025, kimMOFFlowFlowMatching2025, kimFlexibleMOFGeneration2025}, conditioning generation on topology, building blocks, supplied composition, or molecular graphs \citep{kimAtomMOFAllAtomFlow2026, inizanSystemAgenticAI2025}, or using other structured representations \citep{parkMultimodalConditionalDiffusion2025}. Without these restrictions, all-atom MOF generation remains challenging and leads to lower validity than for small crystals \citep{joshiAllatomDiffusionTransformers2025, moreheadZatom1MultimodalFlow2026, renSinAESingleArchitectureFlowMatching2026}. Mofasa narrows this gap with a combination of design choices, including atom-wise latents, residual vector quantization regularization, and a canonical graph ordering \citep{simkusMofasaStepChange2025}, but its validity remains below that of the training structures. Motivated by this challenge, we investigate how the representation used during generation can support larger all-atom crystals.

Generative models in domains such as images or videos typically operate on vector representations with a fixed correspondence, where each vector entry deterministically maps to an image pixel or patch. Atomistic systems, however, are sets without an intrinsic order, such that many different vector representations describe the same physical structure. In particle-space flow matching, a training path connects randomly placed source particles to the atoms of a target crystal. Defining this path requires deciding which source particle moves toward which target atom. We refer to this pairing as the source-to-target correspondence, and the difficulty of learning transport that depends on it the \emph{correspondence problem}. Different pairings describe the same final crystal but prescribe different particle-wise trajectories. The selected pairing defines the training supervision, while the model learns the resulting velocity field without explicit access to the pairing. During generation, no target crystal is available for matching.

In particle-space flow matching with independent coupling, source and target configurations are sampled independently, and particle-wise correspondence is determined implicitly by their indexed representations. Explicit strategies for simplifying this correspondence include canonicalization, which imposes a reproducible ordering in target samples that models can access via an index signal \citep{zhouRethinkingDiffusionModels2026, seongMultimodalCrystalFlow2026, simkusMofasaStepChange2025}, matching-based approaches, which optimize the pairing between source particles and target atoms \citep{kleinEquivariantFlowMatching2023}, or symmetry-reduced representations, which represent symmetry-equivalent atoms through Wyckoff positions or symmetry-inequivalent sites \citep{caoSpaceGroupInformed2025, ekstromkelviniusWyckoffDiffGenerativeDiffusion2025, kazeevWyckoffTransformerGeneration2025}. Stochastic sampling, such as the churn in Crystalite's sampler \citep{veljkovicCrystaliteLightweightTransformer2026}, can additionally separate particles that an equivariant model would otherwise move identically (Appendix~\ref{app:crystalite-churn}). These strategies can improve generation in practice, but none removes the correspondence ambiguity in general: optimal-transport pairings switch discontinuously with the source configuration, and no canonical ordering is continuous over a distribution of point sets \citep{dymEquivariantFramesImpossibility2024}, so nearby structures can receive different orderings. 

In controlled experiments, we find that the training required for reliable generation in index-free permutation-equivariant particle flows increases substantially as set cardinality and target density increase, even when each model is trained on a single target grid. This trend persists when the training pairings are selected using optimal transport. Two properties make this setting hard: optimal-transport pairings switch discontinuously with the source configuration, and a deterministic permutation-equivariant model must give identical predictions to particles with identical states \citep{kabaSymmetryBreakingEquivariant2024, zhangMultisetEquivariantSetPrediction2022}. A learned-slot model can instead assign stable identities to each particle, resulting in fast and reliable learning. These observations motivate removing atom-wise correspondence from generative transport altogether.

This issue is not unique to atomistic generative modeling. Predicting bounding boxes, reconstructing sets, or generating point clouds all require models to produce unordered outputs without assuming a fixed correspondence between prediction and target elements. This has been studied explicitly as the \emph{responsibility problem} in set autoencoding \citep{zhangFSPOOLLEARNINGSET2020}, and methods such as DSPN \citep{zhangDeepSetPrediction2019}, TSPN \citep{kosiorekConditionalSetGeneration2020}, and DETR \citep{carionEndtoEndObjectDetection2020} use invariant representations or permutation-invariant matching to respect this permutation invariance. More generally, permutation-invariant global representations provide a way to describe a set without retaining the identities of its individual elements \citep{zaheerDeepSets2017}. While such representations are not universally lossless and their required capacity depends on the represented set \citep{wagstaffLimitationsRepresentingFunctions2019, wagstaffUniversalApproximationFunctions2022}, they offer a way to remove element-wise correspondence from the representation.

We introduce \textbf{GLASS}---\textbf{G}lobal \textbf{L}atent \textbf{A}ggregation with \textbf{S}lot-based \textbf{S}et Decoding, which encodes each crystal into a permutation-invariant global latent representation and learns its distribution via flow matching. A learned-slot decoder reconstructs the all-atom structure in a single parallel pass. We match predicted and target atoms when training the autoencoder, while flow matching operates entirely in the fixed-dimensional latent space. Concurrent work also applies fixed-dimensional latent flow matching to molecules \citep{yaoFixedDimensionalLatentFlow2026}, using canonical atom sequences and autoregressive decoding. GLASS instead combines a permutation-invariant latent with parallel decoding of the crystal structures.

On MP20, GLASS is competitive with particle-space models, and continued flow training reaches the validity of the training data at every structure size. To establish its scaling to larger structures, our main demonstration is unconditional all-atom MOF generation up to 150 atoms per unit cell, where GLASS approaches the raw structural validity of the QMOF150 training data. On both datasets, flow training moves generation along a validity--novelty frontier.

Our contributions are:

\begin{enumerate}
\item \textbf{The correspondence problem as a scaling bottleneck.} We isolate a failure mode of index-free permutation-equivariant particle flows: even on a single fixed target, training required for reliable generation grows substantially with set size and density under both independent and optimal-transport couplings, whereas breaking the symmetry with learned slots makes the same targets easy to learn at every size.

\item \textbf{Removing correspondence from generative transport.} We introduce GLASS, which removes atom-wise correspondence from generative transport: flow matching operates in a permutation-invariant global latent space, and atom-wise correspondence enters only the reconstruction objective of a slot decoder, without an imposed canonical atom ordering.

\item \textbf{Near-training validity for all-atom MOFs.} GLASS generates unconditional all-atom MOFs with up to 150 atoms at near-training structural validity, demonstrating the scalability of the approach. However, most generated structures recover training samples, revealing a generalization limitation on this dataset.
\end{enumerate}

\section{The Correspondence Problem in Equivariant Set Flows}
\label{sec:correspondence}

Atomistic generative models commonly use permutation-equivariant architectures to respect atom exchangeability. Constructing a particle-space training path then requires pairing source particles with target sites. Since the target set has no intrinsic ordering, many pairings are possible, which each prescribe different particle-wise motions. We study how this ambiguity affects learning as the number of particles grows.

Let $\mathbf{x}_0=(x_{0,1},\ldots,x_{0,N})$ denote the source configuration and $\mathbf{y}=(y_1,\ldots,y_N)$ an ordered representation of the target set. A permutation $\pi\in S_N$ of the $N$ indices pairs source particle $i$ with target site $y_{\pi(i)}$. Writing $\mathbf{y}^\pi=(y_{\pi(1)},\ldots,y_{\pi(N)})$, conditional flow matching \citep{lipmanFlowMatchingGenerative2023,albergoBuildingNormalizingFlows2023} uses
\[
\begin{aligned}
    \mathbf{x}_t^\pi&=(1-t)\mathbf{x}_0+t\mathbf{y}^\pi,
    \qquad \mathbf{u}^\pi=\mathbf{y}^\pi-\mathbf{x}_0,\\
    \mathcal{L}_{\mathrm{CFM}}&=\mathbb{E}_{t,\mathbf{x}_0,\mathbf{y},\pi}\!\left[\left\|v_\theta(\mathbf{x}_t^\pi,t)-\mathbf{u}^\pi\right\|_F^2\right].
\end{aligned}
\]
In our fixed-grid experiment, independent coupling pairs source particles with randomly ordered target sites. Optimal transport (OT) instead selects $\pi^\star\in\arg\min_{\pi\in S_N}\sum_i\|x_{0,i}-y_{\pi(i)}\|_2^2$ \citep{kleinEquivariantFlowMatching2023}. Under squared loss, the optimal predictor is the conditional mean of the prescribed velocity given the intermediate state and time, which the model learns without explicit access to the permutation. During generation, no target is supplied for matching, and the learned velocity field is integrated to generate the final state. 

\begin{figure}[t]
    \centering
    \includegraphics[width=\linewidth]{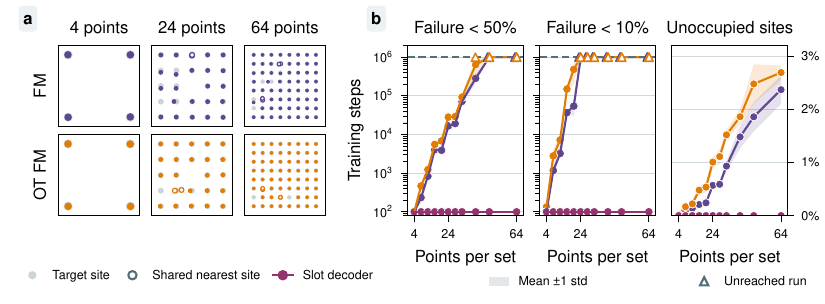}
    \caption{\textbf{Training requirements and site coverage in index-free permutation-equivariant particle flows.}
    \textbf{a)} Target grids and generated samples from independent and optimal-transport (OT) couplings. Hollow markers indicate generated particles sharing a nearest target site.
    \textbf{b)} Left and center: training steps to whole-set failure rates below $50\,\%$ and $10\,\%$. Right: mean unoccupied-site fraction at final checkpoints. The slot decoder realizes each target from learned slots with Hungarian matching and reaches every threshold at its first evaluation. Curves summarize three runs; shading denotes one standard deviation. Hollow triangles mark unreached thresholds.}
    \label{fig:toy}
\end{figure}

\subsection{Cardinality and Generation Reliability}

To separate the effect of cardinality from that of learning a data distribution, we reduce generation to its simplest form: reproducing a single fixed set of sites. For each $N$, we train a permutation-equivariant Transformer \citep{vaswaniAttentionAllYou2017} on one fixed target grid in $[-1,1]^2$ (Figure~\ref{fig:toy}a). Particles carry only positions, with no indexing signal. Cardinality and density increase together, requiring finer spatial resolution. We assign each generated particle to its nearest target site; a sample fails if two particles select the same site (\emph{duplicate-site occupancy}).

Figure~\ref{fig:toy}b reports training steps required to reach whole-set failure rates below $50\,\%$ and $10\,\%$, and the final unoccupied-site fraction. Both couplings require more training at larger $N$, with several runs missing the thresholds within 1M training steps; optimal-transport coupling does not remove this dependence (Table~\ref{tab:sampling-resolution}, Appendix~\ref{app:toy-experiments}). As a control, we train the same targets with a latent-free version of the GLASS decoder: learned slot embeddings processed by self-attention layers of comparable size and trained with Hungarian matching to the target sites. This model reaches both thresholds at the first evaluation at 100 gradient steps for every $N$ and leaves no site unoccupied (Figure~\ref{fig:toy}b; worst generated outputs shown in Figure~\ref{fig:toy-decoder}).

\subsection{Assignment Sensitivity and Equivariance}

As the difficulty appears even for a single fixed target, we examine two properties of the transport problem itself that become more demanding as sets grow larger and denser. First, small changes in source positions can switch the OT pairing and prescribe sharply different initial velocities. For example, consider target sites $y_1=(-1,0)$, $y_2=(1,0)$ and source particles at $x_{0,1}=(\varepsilon,1)$, $x_{0,2}=(-\varepsilon,-1)$. Changing the sign of $\varepsilon$ switches the first particle's destination between the two sites. The model must learn this discontinuous dependence across source configurations, even though each prescribed path is continuous in time. While exact OT paths remain separated for distinct target sites and their intermediate configurations determine the pairing (Appendix~\ref{app:assignment-boundary}), learning this discrete boundary can be challenging.

Second, permutation equivariance imposes a separate constraint: a deterministic equivariant predictor must give identical outputs to particles with identical states \citep{zhangMultisetEquivariantSetPrediction2022,kabaSymmetryBreakingEquivariant2024}, and separating nearby particles requires correspondingly high sensitivity (Appendix~\ref{app:equivariance-sensitivity}). Denser targets tighten both constraints, since smaller positional errors change the nearest target site.

For a single target, learned slot identities break this symmetry directly (Figure~\ref{fig:toy}b), making learning on these samples trivial. When considering a distribution of crystals, no continuous canonical ordering exists \citep{dymEquivariantFramesImpossibility2024} to assign a stable ordering. GLASS instead learns slots to decode structures and performs generation in a permutation-invariant latent space, decoupling the correspondence problem from the generative transport.

\section{GLASS Architecture}
\label{sec:glass}

\begin{figure}[t]
    \centering
    \includegraphics[width=\linewidth]{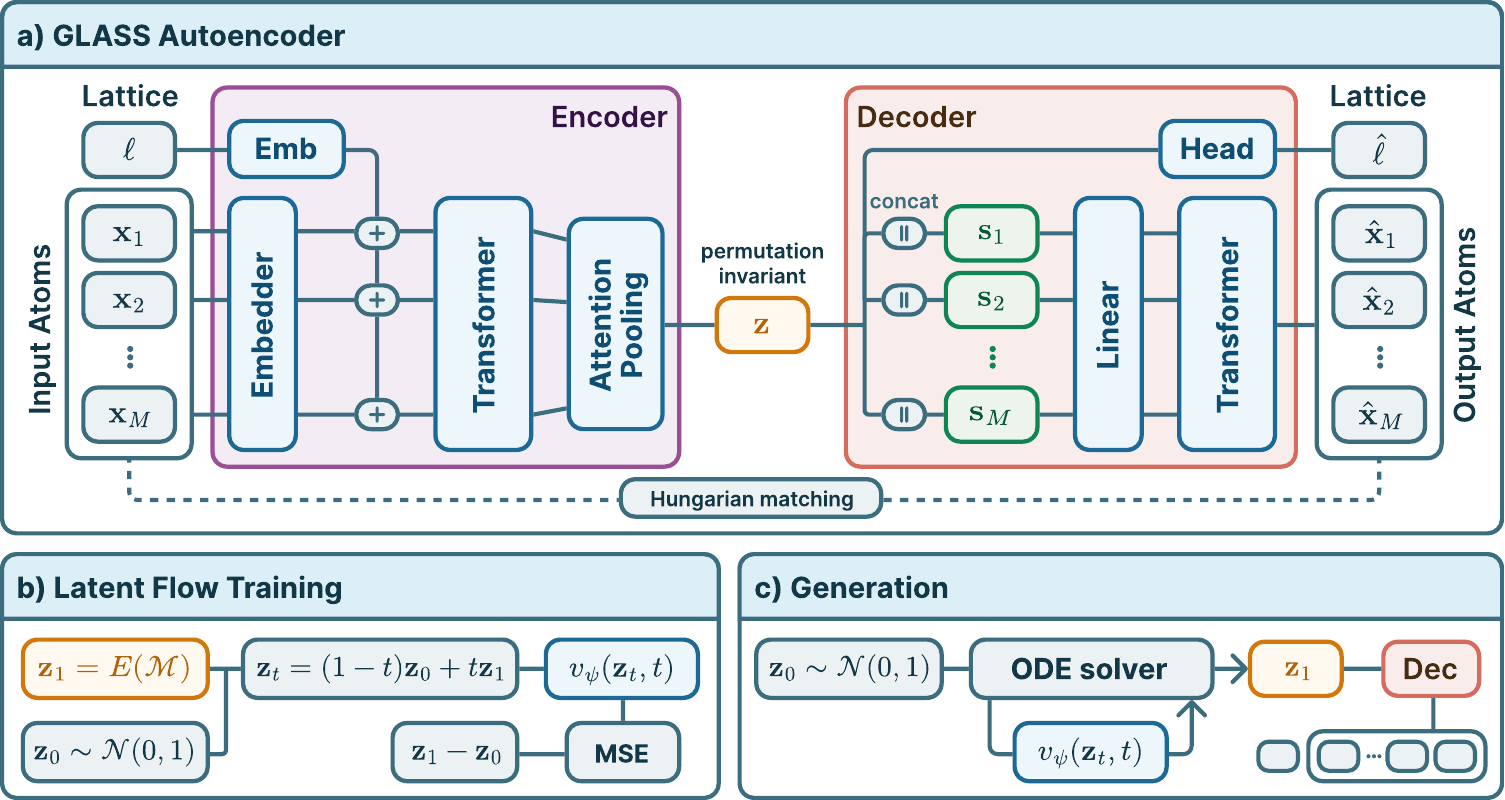}
    \caption{\textbf{GLASS architecture.} A Transformer encoder and invariant attention pooling map a crystal to a global latent vector. Learned decoder slots predict atom types and coordinates in parallel, while a separate head predicts the lattice. Flow matching generates new latents, which are decoded into all-atom structures.}
    \label{fig:architecture}
\end{figure}

Building on Section~\ref{sec:correspondence}, GLASS removes the correspondence problem from generative transport by separating transport from atom-wise reconstruction. An autoencoder encodes each crystal into a permutation-invariant global latent vector $\mathbf{z}\in\mathbb{R}^{d_z}$ and reconstructs it with a learned-slot decoder. We then train a flow in this fixed-dimensional latent space (Figure~\ref{fig:architecture}).

\subsection{Autoencoder}

A crystal structure $\mathcal{M}=(\mathbf{A},\mathbf{F},\mathbf{L})$ with $n$ atoms is described by three variables: atom types $\mathbf{A}\in\{1,\ldots,94\}^n$, represented by atomic numbers; fractional coordinates $\mathbf{F}\in[0,1)^{n\times3}$ with rows $\mathbf{f}_i^\top$; and lattice parameters $\mathbf{L}=(a,b,c,\alpha,\beta,\gamma)^\top$ with lengths $a,b,c$ and angles $\alpha,\beta,\gamma$. We denote the unordered atom set by $x=\{\mathbf{x}_1,\ldots,\mathbf{x}_n\}$, where $\mathbf{x}_i=(A_i,\mathbf{f}_i^\top)^\top$ is the vector describing atom $i$.

We adopt a tokenization that assigns one token to each atom and adds a lattice embedding onto every token. Atom types are represented through learned embeddings, while fractional coordinates are embedded using periodic Fourier features, making the periodicity directly visible to the model. We represent the lattice as $\boldsymbol{\ell}=[\log a,\log b,\log c,\cos\alpha,\cos\beta,\cos\gamma]^\top$, which is embedded separately and added to each atom token, giving
\begin{equation}
\mathbf{t}_i^{(0)}
=
E_A(A_i)
+
E_f(\mathbf{f}_i)
+
E_L(\boldsymbol{\ell}).
\end{equation}
The resulting token set is processed by multiple Transformer encoder layers before a final attention-pooling operation with a global learned query. Throughout the encoder, permutation equivariance is preserved, while the final pooled latent is invariant to permutation.

The pooled bottleneck breaks the input-to-output atom correspondence. To decode a structure from the global latent code, we use $M$ learned slot embeddings $\mathbf{s}_j$, each of which is concatenated with the latent representation and linearly projected,
\begin{equation}
\mathbf{u}_j=\mathbf{W}_u[\mathbf{s}_j;\mathbf{z}]+\mathbf{b}_u.
\end{equation}
The resulting slot representations are passed through Transformer layers and decoded through separate coordinate and type heads. The learned slot identities provide a stable symmetry-breaking signal, allowing the model to produce distinct output atoms without retaining the original ordering of the input atoms. The type head predicts atom types and an additional empty class for masked atoms. Each slot is assigned its most probable type; empty slots are discarded, determining the output atom count up to $M$. A separate head predicts the lattice directly from the global latent. The decoder thus generates atom count, types, coordinates, and lattice together. Representation and padding details are given in Appendix~\ref{app:glass-details}.

The autoencoder is trained to reconstruct its inputs. Since the global bottleneck does not retain the input ordering, we match predicted and target atoms using the Hungarian algorithm \citep{kuhnHungarianMethodAssignment1955}. For a structure with $n$ atoms, padded to $M$ entries, the periodic squared Cartesian error between the position predicted by slot $i$ and target atom $j$ is
\begin{equation}
\delta^2_{ij}
=
\min_{\mathbf{k}\in\{-1,0,1\}^3}
\left(\mathbf{f}_j-\hat{\mathbf{f}}_i-\mathbf{k}\right)^\top
\mathbf{G}
\left(\mathbf{f}_j-\hat{\mathbf{f}}_i-\mathbf{k}\right),
\end{equation}
where $\mathbf{G}$ is the Gram matrix of the target cell. With $\hat A_{i,a}$ the predicted probability of type $a$ for slot $i$, and type $0$ marking empty slots and target padding, the matching cost and assignment are
\begin{equation}
    C_{ij}=-\frac{\lambda_A}{M}\log\hat A_{i,A_j}
    +\frac{\lambda_f}{3n}\mathbf{1}[A_j\neq0]\,\delta^2_{ij},
    \qquad
    \pi^\star=\arg\min_{\pi\in S_M}\sum_{i=1}^M C_{i,\pi(i)}.
\end{equation}
Given $\pi^\star$, the type and coordinate losses are
\begin{equation}
\mathcal L_{\mathrm{type}}
=
-\frac{1}{M}
\sum_{i=1}^{M}
\log \hat A_{i,A_{\pi^\star(i)}},
\qquad
\mathcal L_{\mathrm{coord}}
=
\frac{1}{3n}
\sum_{i=1}^{M}
\mathbf{1}[A_{\pi^\star(i)}\neq0]\,
\delta^2_{i,\pi^\star(i)},
\end{equation}
and the lattice head is trained with $\mathcal{L}_{\mathrm{cell}}=\frac{1}{6}\sum_{k}(\hat\ell_k-\ell_k)^2$.

To improve local geometry reconstruction, we add an auxiliary short-range pair-distance loss. For every pair of real atoms under $\pi^\star$, we compare predicted and target minimum-image distances $\hat d_{ij}$ and $d_{ij}$ through the relative error $e_{ij}=((\hat d_{ij}-d_{ij})/d_{ij})^2$,
\begin{equation}
\mathcal{L}_{\mathrm{pair}}
=
\frac{\sum_{i<j}w(d_{ij})\,e_{ij}}{\sum_{i<j}w(d_{ij})},\qquad
w(d)
=
\begin{cases}
1, & d\leq3,\\
\frac{1}{2}\left[1+\cos\left(\pi\frac{d-3}{2}\right)\right], & 3<d<5,\\
0, & d\geq5,
\end{cases}
\end{equation}
The total objective is $\mathcal L_{\mathrm{AE}}=\lambda_A\mathcal L_{\mathrm{type}}+\lambda_f\mathcal L_{\mathrm{coord}}+\lambda_{\mathrm{cell}}\mathcal L_{\mathrm{cell}}+\lambda_{\mathrm{pair}}\mathcal L_{\mathrm{pair}}$.

\subsection{Latent Flow Matching}

After training the autoencoder, we freeze its parameters and train a time-conditioned residual multilayer perceptron as the latent flow model. Given an encoded data latent $\mathbf{z}_1$ and Gaussian noise $\mathbf{z}_0$, we interpolate as $\mathbf{z}_t=(1-t)\mathbf{z}_0+t \mathbf{z}_1$ and train a vector field $v_\psi(\mathbf{z}_t,t)$ to predict the constant velocity $\mathbf{z}_1-\mathbf{z}_0$. The flow-matching objective is
\begin{equation}
\mathcal L_{\mathrm{flow}}
=
\mathbb E_{\mathbf{z}_0,\mathbf{z}_1,t}
\left[
\frac{1}{d_z}
\left\|
v_\psi(\mathbf{z}_t,t)-(\mathbf{z}_1-\mathbf{z}_0)
\right\|_2^2
\right].
\end{equation}

For sample generation, the learned flow is integrated from Gaussian noise in latent space. The generated latent is transformed back to the original latent scale and decoded into the full crystal structure in a single parallel pass. Sampling settings are given in Appendix~\ref{sec:model_settings}.

For $K$ vector-field evaluations, generation costs $T_{\mathrm{GLASS}}=K\,C_{\mathrm{latent}}(d_z)+C_{\mathrm{dec}}(M)$. At fixed model capacity, latent integration is independent of atom count, and the quadratic attention over the $M$ decoder slots is evaluated only once per generated structure (Appendix~\ref{sec:computational_cost}).

\providecommand{\glassfit}[1]{%
  \begingroup
  \sbox{0}{#1}%
  \ifdim\wd0>\linewidth
    \resizebox{\linewidth}{!}{\usebox{0}}%
  \else
    \usebox{0}%
  \fi
  \endgroup
}

\section{Experiments}
\label{sec:mp20-experiments}

We first evaluate autoencoder generalization, generated structural validity, and novelty on small crystals, before investigating the model on larger all-atom MOFs.

\begin{figure}[!t]
    \centering
    \includegraphics[width=\linewidth]{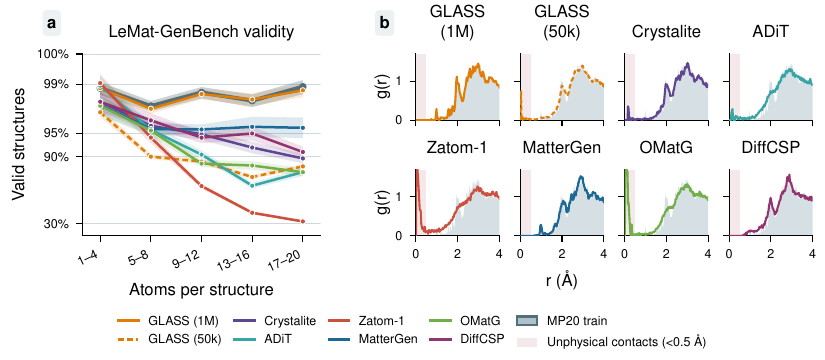}
    \caption{\textbf{Validity by size and local geometric fidelity on MP20-generated structures.}
    \textbf{(a)} LeMat-GenBench validity before relaxation as a function of atom count, including GLASS at 50k and 1M flow-training steps, each with $10,000$ samples. Bands show 95\% Wilson intervals; the vertical axis uses an arcsinh scale that expands differences near $100\,\%$.
    \textbf{(b)} Radial distribution functions for 13--16 atoms, compared with MP20 training data (grey). The red shaded interval marks interatomic distances below $0.5\,\text{\AA}$.}
    \label{fig:mp20_validity}
\end{figure}

We train GLASS on MP20, containing Materials Project crystals with up to 20 atoms \citep{xieCrystalDiffusionVariational2022,jainCommentaryMaterialsProject2013} and compare generation performance with Crystalite \citep{veljkovicCrystaliteLightweightTransformer2026}, ADiT \citep{joshiAllatomDiffusionTransformers2025}, Zatom-1 \citep{moreheadZatom1MultimodalFlow2026}, MatterGen \citep{zeniGenerativeModelInorganic2025}, OMatG \citep{hollmerOpenMaterialsGeneration2025}, and DiffCSP \citep{jiaoCrystalStructurePrediction2023}, and include LeMat-GenBench leaderboard results for OMatG-FC \citep{martirossyanFrameworkConstrainedMaterialsGeneration2026}, MiAD \citep{okhotinMiADMirageAtom2026}, and Chemeleon-2 \citep{parkGuidingGenerativeModels2026} in Table 1.

On this dataset, the autoencoder shows a validation gap: held-out reconstruction RMSD is \MpValCellRmsd\,\AA, against \MpTrainCellRmsd\,\AA\ on training structures. Of 512 held-out reconstructions, $81.64\,\%$ pass LeMat validity checks, compared with $97.85\,\%$ of their targets.

Despite this reconstruction gap, generated structures can closely match training-set validity, based on LeMat-GenBench's validity criteria \citep{betalaLeMatGenBenchUnifiedEvaluation2025}. After 1M flow steps, $30,000$ samples from three training seeds reach $98.38\,\%$ raw validity, against $98.45\,\%$ for training data, and stay within $0.2$ percentage points in every size bin (Figure~\ref{fig:mp20_validity}a; Table~\ref{tab:mp20-lemat-validity}). Every comparison model loses validity between the smallest and largest bins, most strongly for Zatom-1, from $99.04\,\%$ at 1--4 atoms to $34.17\,\%$ at 17--20 atoms. In contrast, MatterGen's decline is modest, and we show more details in Appendix~\ref{app:mp20_validity_full}. These trends are consistent with the correspondence difficulty shown in Section~\ref{sec:correspondence}.

Radial distribution functions show how this agreement extends to local geometry (Figure~\ref{fig:mp20_validity}b). GLASS at 1M closely matches the training peaks, while comparison models show broader or distorted peaks and excess short-distance weight, particularly unphysical close contacts for Zatom-1 and OMatG. GLASS at 50k captures the main distribution but retains geometric defects.

However, matching training geometry does not establish novelty or energetic stability, which are important targets for materials discovery. We therefore evaluate GLASS with LeMat-GenBench (Table~\ref{tab:lemat-genbench}): SUN and MSUN measure unique, novel yields in disjoint stable and metastable MLIP energy ranges, with novelty assessed against the benchmark's broad reference database. For models with relaxed generated samples, validity nearly saturates at $96$--$97\,\%$ across the displayed methods; novelty and energetic stability remain distinguishing factors.

For GLASS, these criteria favor an earlier checkpoint: longer flow training improves validity but reduces valid-and-novel yield from $45.45\,\%$ at 50k to $3.16\,\%$ at 1M (Figure~\ref{fig:checkpoint_tradeoff}). We select the 50k checkpoint for a balance between validity and novelty. With relaxation using NequIP-OAM-L \citep{batznerE3equivariantGraphNeural2022, kavanaghNequIPAllegroFoundation2026}, GLASS samples reach $96.56\,\%$ validity, $52.71\,\%$ novelty, and the highest metastable fraction in the displayed group. A SUN score of $1.92\,\%$ leads this group, and MSUN ($21.16\,\%$) is close to Crystalite ($22.60\,\%$). GLASS is thus competitive on MP20, although its raw validity on the 50k checkpoint is lower than the unrelaxed baselines. Protocols and complementary Crystalite scores are shown in Appendix~\ref{sec:experimental_details} and Table~\ref{tab:crystalite-dng}.

\begin{table*}[t]
\centering
\caption{\textbf{LeMat-GenBench evaluation on MP20.} Results are grouped by input pre-relaxation. GLASS uses the \MpTableCheckpoint{} flow checkpoint; reference values are obtained from the official HuggingFace space \citep{betalaLeMatGenBenchUnifiedEvaluation2025}. Additional models appear in Table~\ref{tab:lemat-genbench-full}. Stability metrics are MLIP-based estimates, and MSUN excludes SUN. Bold denotes the best displayed mean within each group.\MpTableNote}
\label{tab:lemat-genbench}
\begingroup
\small
\setlength{\tabcolsep}{4pt}
\renewcommand{\arraystretch}{1}
\glassfit{%
\begin{tabular}{lrrrrrrrrr}
\toprule
Model & Valid & Unique & Novel & Stable & Metastable & SUN & MSUN & E above hull & Relax. RMSD \\
& (\%) $\uparrow$ & (\%) $\uparrow$ & (\%) $\uparrow$ & (\%) $\uparrow$ & (\%) $\uparrow$ & (\%) $\uparrow$ & (\%) $\uparrow$ & (eV/atom) $\downarrow$ & (\AA) $\downarrow$ \\
\midrule
\multicolumn{10}{c}{\textbf{Pre-relaxed inputs}} \\
\midrule
Crystalite & \textbf{97.20} & 95.80 & 53.20 & 12.70 & 51.60 & 1.50 & \textbf{22.60} & 0.0905 & 0.1322 \\
OMatG & 96.40 & 95.20 & 51.20 & 11.60 & 49.80 & 1.00 & 18.00 & 0.0956 & 0.0759 \\
MiAD & 96.20 & 94.30 & 40.20 & 6.40 & 62.00 & 1.00 & 16.60 & 0.0804 & 0.2494 \\
MatterGen & 95.70 & 95.10 & \textbf{70.50} & 2.00 & 33.40 & 0.20 & 15.00 & 0.1834 & 0.3878 \\
OMatG-FC & \textbf{97.20} & 92.80 & 28.90 & \textbf{18.40} & 57.90 & 1.70 & 12.00 & \textbf{0.0694} & \textbf{0.0685} \\
\midrule
GLASS + NequIP & 96.56 & 95.48 & 52.71 & 11.75 & \textbf{63.08} & \textbf{1.92} & 21.16 & 0.1261 & 0.1258 \\
 & $\pm$ 0.39 & $\pm$ 0.66 & $\pm$ 1.40 & $\pm$ 0.51 & $\pm$ 1.52 & $\pm$ 0.18 & $\pm$ 1.40 & $\pm$ 0.0080 & $\pm$ 0.0071 \\
\midrule
\multicolumn{10}{c}{\textbf{Inputs without pre-relaxation}} \\
\midrule
Chemeleon2 & 95.20 & 88.10 & \textbf{71.60} & 0.00 & 39.80 & 0.00 & \textbf{21.20} & \textbf{0.1557} & \textbf{0.4226} \\
DiffCSP & \textbf{95.70} & \textbf{94.80} & 66.20 & 2.30 & 29.80 & 0.10 & 8.50 & 0.2747 & 0.5857 \\
\midrule
GLASS & 89.29 & 88.32 & 45.65 & 0.93 & \textbf{41.95} & 0.15 & 11.49 & 0.3140 & 0.4290 \\
 & $\pm$ 0.59 & $\pm$ 0.68 & $\pm$ 1.20 & $\pm$ 0.10 & $\pm$ 1.65 & $\pm$ 0.06 & $\pm$ 0.61 & $\pm$ 0.0083 & $\pm$ 0.0077 \\
\bottomrule
\end{tabular}
}
\endgroup
\end{table*}

\section{High-Validity Generation of Metal--Organic Frameworks}
\label{sec:mof}

To test whether this structural fidelity extends to larger crystals, we apply GLASS to QMOF \citep{rosenMachineLearningQuantumchemical2021}, with up to 150 atoms per cell, using the same architecture and a 64-dimensional latent. Generation is unconditioned on building blocks, topology, or composition; validity is assessed with MOFChecker \citep{jinMOFCheckerPackageValidating2025}.

We find that the autoencoder gap is larger here: held-out RMSD is \QmofValCellRmsd\,\AA, against \QmofTrainCellRmsd\,\AA\ on training structures. Increasing latent dimension from 32 to 128 barely changes this gap (Appendix~\ref{sec:qmof_capacity}).

As on MP20, the model can achieve high generation validity, which rises through our reported 1M checkpoint (Figure~\ref{fig:qmof-checkpoint-tradeoff}). GLASS reaches \QmofValidity\,\% raw validity, against \QmofTrainingValidity\,\% for training data, $52.9\,\%$ for Mofasa \citep{simkusMofasaStepChange2025}, and $15.1$--$16.3\,\%$ for ADiT, Zatom-1, and SinAE \citep{joshiAllatomDiffusionTransformers2025,moreheadZatom1MultimodalFlow2026,renSinAESingleArchitectureFlowMatching2026}.

\begin{figure*}[t]
    \centering
    \begin{minipage}[t]{0.38\textwidth}
        \vspace{0pt}
        \centering
        \textbf{(a) Structural validity by size}
        \vspace{2pt}
        \includegraphics[width=\linewidth]{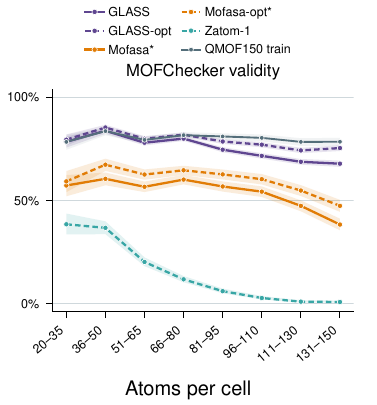}
    \end{minipage}
    \hfill
    \begin{minipage}[t]{0.6\textwidth}
        \vspace{0pt}
        \centering
        \textbf{(b) Raw MOFChecker validity breakdown}
        \vspace{4pt}
        \scriptsize
        \setlength{\tabcolsep}{2.2pt}
        \glassfit{%
\begin{tabular}{@{}lrrrrrr@{}}
            \toprule
            Criterion & Train & GLASS & Mofasa$^*$ & ADiT & Zatom-1 & SinAE \\
            \midrule
\textbf{Overall valid} $\uparrow$ & 80.3 & \textbf{74.8} & 52.9 & 15.7 & 15.1 & 16.3 \\
Has C $\uparrow$ & 100.0 & 99.1 & 98.4 & \textbf{100.0} & \textbf{100.0} & \textbf{100.0} \\
Has H $\uparrow$ & 99.8 & 98.9 & 98.4 & 99.6 & \textbf{100.0} & 99.8 \\
Atom overlap $\downarrow$ & 0.0 & \textbf{1.3} & 2.8 & 8.3 & 8.8 & 5.2 \\
Overcoord. C $\downarrow$ & 0.0 & \textbf{0.5} & 2.5 & 23.6 & 1.1 & 12.8 \\
Overcoord. N $\downarrow$ & 0.0 & 0.1 & 1.6 & 1.5 & \textbf{0.0} & 0.4 \\
Overcoord. H $\downarrow$ & 0.0 & \textbf{1.0} & 2.4 & \textbf{1.0} & 2.9 & 1.3 \\
Undercoord. C $\downarrow$ & 5.9 & \textbf{8.2} & 20.1 & 60.0 & 65.4 & 39.2 \\
Undercoord. N $\downarrow$ & 6.9 & \textbf{11.4} & 12.9 & 39.1 & 22.8 & 26.9 \\
Undercoord. rare earth $\downarrow$ & 0.0 & 0.1 & 2.1 & 0.4 & \textbf{0.0} & 0.2 \\
Has metal $\uparrow$ & 100.0 & 99.0 & 97.7 & \textbf{100.0} & \textbf{100.0} & \textbf{100.0} \\
Lone molecule $\downarrow$ & 9.7 & \textbf{12.1} & 31.3 & 72.9 & 80.5 & 49.8 \\
High charge $\downarrow$ & 1.0 & 1.4 & 3.5 & 0.9 & 0.5 & \textbf{0.2} \\
Terminal oxo $\downarrow$ & 0.0 & 0.5 & 2.5 & 2.6 & \textbf{0.3} & 3.1 \\
Undercoord. alk./alk.-earth $\downarrow$ & 0.2 & \textbf{0.3} & 3.0 & 1.0 & \textbf{0.3} & 0.7 \\
Geom. exposed metal $\downarrow$ & 1.7 & 2.6 & 5.3 & 7.0 & \textbf{1.8} & 4.0 \\
            \bottomrule
        \end{tabular}%
        }
    \end{minipage}
    \caption{\textbf{Structural validity of raw and relaxed all-atom MOF generations.}
    GLASS uses the 1M-step flow checkpoint.
    \textbf{(a)} MOFChecker validity by atom count before and after relaxation (opt), compared with QMOF150 training data; bands show 95\% Wilson intervals. ADiT and SinAE are absent because no released QMOF samples or checkpoints were available.
    \textbf{(b)} Raw-generation validity and individual checks (\%). Arrows indicate the preferred direction; defect flags can overlap.
    $^*$Mofasa: precomputed MOFChecker results for 10,000 released samples with 20--150 atoms and their relaxed counterparts; trained on QMOF structures up to 170 atoms with a different split. All models are shown in Appendix~\ref{sec:qmof_additional}.}
    \label{fig:qmof-results}
\end{figure*}

This advantage persists across sizes (Figure~\ref{fig:qmof-results}a). GLASS nearly matches training validity at 20--35 atoms and retains $\QmofLargeValidity\,\%$ at 131--150 atoms, against $\QmofTrainingLargeValidity\,\%$ for training data, $\MofasaLargeValidity\,\%$ for Mofasa, and $\ZatomLargeValidity\,\%$ for Zatom-1. Its lower rates of disconnected molecules and undercoordination explain much of the advantage over ADiT, Zatom-1, and SinAE (Figure~\ref{fig:qmof-results}b). Relaxation with eSEN-OAM+D3 \citep{fuLearningSmoothExpressive2025, grimmeConsistentAccurateInitio2010} raises overall validity to \QmofRelaxedValidity\,\% (Appendix~\ref{sec:qmof_additional}).

This structural fidelity, however, comes with limited novelty. Under Mofasa's MOFid \citep{buciorIdentificationSchemesMetal2019} convention, valid, novel, and unique yield (VNU) is \QmofVnu\,\%, versus Mofasa's reported $42.4\,\%$. However, different training splits make this comparison only indicative (Appendix~\ref{sec:qmof_mofid_protocol}). Novelty analysis with StructureMatcher \citep{ongPythonMaterialsGenomics2013} finds that \QmofTrainingMatchRate\,\% of valid GLASS samples match training structures, compared to only \MofasaAnySplitMatchRate\,\% of valid Mofasa samples matching any QMOF150 structure (Appendix~\ref{sec:pipeline_localization}). GLASS's validity therefore primarily reflects reliable generation of familiar frameworks.

Unlike on MP20, earlier checkpoints offer little improvement: valid, unmatched yield falls from \QmofEarlyValidUnmatchedYield\,\% at 250k to \QmofValidUnmatchedYield\,\% at 1M. This brings the larger autoencoder validation gap into focus: its poor generalization may restrict validity before the flow closely fits the training latents. Decoding away from training encodings may fail, favoring familiar frameworks at the later flow checkpoints. 

To test whether the autoencoder gap can be closed, we train the GLASS autoencoder on LeMat-Bulk, with 5M training structures of up to 20 atoms, where held-out reconstruction nearly matches training ($\LematBulkValCellRmsd$ versus $\LematBulkTrainCellRmsd$\,\AA; Appendix~\ref{app:lemat-bulk}). Thus, the small-crystal validation gap is largely closed in this setting, supporting that broader autoencoder pretraining can improve generalization.

\section{Discussion and Conclusion}
\label{sec:discussion}

We introduced GLASS to separate the correspondence problem from generative transport. We showed that GLASS is competitive on small-crystal generation and achieves leading structural validity among the compared all-atom MOF models, but novel framework generation remains an unresolved limitation. On MP20, the 50k flow checkpoint combines high validity after relaxation with SUN and MSUN yields near the top of the comparison; at 1M, raw validity and local geometry closely match the training data, at the cost of novelty. On QMOF150, GLASS generates structures with up to 150 atoms at validity approaching that of the training data, without building blocks, topology, or composition as input. Most valid outputs, however, reproduce known training frameworks. 

The validity results support separating generative transport from atom correspondence. Our fixed-grid experiments isolate a mechanism consistent with the size-dependent declines of several all-atom models: index-free permutation-equivariant flows need more training to generate a fixed target as the number of particles grows, while a slot decoder with learned identities realizes the same targets within a small, constant budget at every size. GLASS removes this correspondence problem from transport by learning the flow in a permutation-invariant global latent space. The evaluations show that this representation can preserve local geometry and framework connectivity across substantially larger atomistic structures.

A central remaining difficulty is decoding unseen structures reliably. Universal continuous invariant representations of arbitrary sets require sufficient latent dimension as cardinality grows \citep{wagstaffLimitationsRepresentingFunctions2019}. Yet 32 latent dimensions already fit the QMOF150 training structures closely, and increasing this dimension to 128 changes reconstruction errors only modestly (Appendix~\ref{sec:qmof_capacity}). The much smaller train--validation gap on LeMat-Bulk suggests that this gap is not intrinsic to the architecture, although the comparison does not isolate dataset size. Whether pretraining on larger MOF datasets improves reconstruction and valid novel yield remains to be tested.

A global latent also enables new future directions: property-informed latent training, as explored in Crys-JEPA \citep{liuCrysJEPAAcceleratingCrystal2026} and EF-TALFM \citep{yaoFixedDimensionalLatentFlow2026}, could additionally favor energetically useful or property-targeted regions. 

\paragraph{Limitations.}
High MOF validity on QMOF150 comes with little novelty, and a validity deficit remains for the largest frameworks. MOFChecker tests selected structural and chemical criteria; passing these checks does not establish energetic stability. The MP20 stability results are MLIP-based estimates and can differ from first-principles evaluations. The immediate priority for future work is improved autoencoder generalization and measuring whether this increases the yield of valid, novel frameworks.

\subsection*{AI use statement}

In this work, we used LLMs to assist with software implementation, debugging, and code review; to provide feedback on experimental methodology and implementation; and to provide feedback on and assist in checking mathematical arguments. LLMs were also used for literature retrieval and discovery, and to suggest, draft, and improve formulations in the manuscript.

All AI-assisted code, mathematical arguments, citations, and manuscript text were critically reviewed and verified by the authors. The scientific questions, methodological choices, experiments, interpretation of results, and final claims were determined by the authors. LLMs were not used to generate synthetic datasets. The authors take full responsibility for the final content of this work.




\subsection*{Reproducibility statement}

We provide code for training, sampling, and evaluation at \url{https://github.com/henk789/glass}. Appendix D specifies datasets and splits, autoencoder and flow settings (Tables 3 and 4), reconstruction metrics, geometry relaxation, and the LeMat-GenBench and Crystalite evaluation protocols. MOF validity uses MOFChecker 0.9.6, and similarity to training structures uses the StructureMatcher settings provided in Appendix G.3. The fixed-grid experiment and slot-decoder control are specified in Appendix C, with code contained in the released repository. Main results report means and sample standard deviations over three training runs; sampling seeds and sample counts are given with each evaluation. Baseline samples are taken from public releases or generated from released checkpoints (Appendix F.3).


\subsubsection*{Acknowledgments}
The authors thank the International Max Planck Research School for Intelligent Systems (IMPRS-IS) for supporting Hendrik Kra\ss. This work was funded by the Deutsche Forschungsgemeinschaft (DFG, German Research Foundation) -- Project number 569019417. The authors gratefully acknowledge the computing time made available to them on the high-performance computer at the NHR Center of TU Dresden. This center is jointly supported by the Federal Ministry of Research, Technology and Space of Germany and the state governments participating in the NHR (www.nhr-verein.de/unsere-partner). SMM acknowledges support from the University of Toronto's Data Science Institute and the Acceleration Consortium, which receives funding from the Canada First Research Excellence Fund (CFREF).

\bibliography{iclr2027_conference}
\bibliographystyle{iclr2027_conference}

\appendix
\raggedbottom

\FloatBarrier
\section{Related Work}
\label{app:related-work}

\paragraph{Molecular and Crystal Generative Models.}
Direct atomistic models retain particle-indexed coordinates and types throughout generation. Examples include EDM \citep{hoogeboomEquivariantDiffusionMolecule2022} and SemlaFlow \citep{irwinSemlaFlowEfficient3D2025} for molecules, and DiffCSP \citep{jiaoCrystalStructurePrediction2023}, FlowMM \citep{millerFlowMMGeneratingMaterials2024}, OMatG \citep{hollmerOpenMaterialsGeneration2025}, CrystalFlow \citep{luoCrystalFlowFlowbasedGenerative2025}, Crystalite \citep{veljkovicCrystaliteLightweightTransformer2026}, and MatterGen \citep{zeniGenerativeModelInorganic2025} for crystals.

Latent models can also retain element-wise structure: GeoLDM \citep{xuGeometricLatentDiffusion2023} uses point-structured latents; ADiT \citep{joshiAllatomDiffusionTransformers2025} and UAE-3D \citep{luoUnifiedLosslessLatent2025} use atom- or token-structured representations; and SinAE \citep{renSinAESingleArchitectureFlowMatching2026} uses variable-length atomic latents and an iterative flow-matching decoder. Mofasa \citep{simkusMofasaStepChange2025} combines global and atom-wise latents. CDVAE \citep{xieCrystalDiffusionVariational2022} encodes a crystal globally, but conditions annealed Langevin dynamics on randomly initialized atoms. Fourier Transformers \citep{duerschFourierTransformersLatent2026} instead generate reciprocal-space densities. EF-TALFM \citep{yaoFixedDimensionalLatentFlow2026} transports a single molecule-level latent and reconstructs molecules autoregressively from a canonical atom sequence. Canonical ordering is also used by CanonFlow \citep{zhouRethinkingDiffusionModels2026}, MCFlow \citep{seongMultimodalCrystalFlow2026}, and Mofasa.

\paragraph{Generation of Large MOFs.}
MOFDiff \citep{fuMOFDiffCoarsegrainedDiffusion2024} generates coarse-grained building blocks. MOFFlow \citep{kimMOFFlowFlowMatching2025} and MOF-BFN \citep{jiaoMOFBFNMetalOrganicFrameworks2025} predict lattice parameters and rigid-body placements of supplied blocks; MOFFlow-2 \citep{kimFlexibleMOFGeneration2025} extends this to generated blocks and torsional degrees of freedom. BBA MOF Diffusion \citep{duanBuildingBlockAwareGenerative2025} uses building-block and topology representations, while MOFFUSION \citep{parkMultimodalConditionalDiffusion2025} maps generated pore representations to framework components. AtomMOF \citep{kimAtomMOFAllAtomFlow2026} conditions on building-block graphs, and MOFGen \citep{inizanSystemAgenticAI2025} uses composition-conditioned generation within a refinement and synthesis pipeline. Mofasa jointly generates atom types, coordinates, and lattices for structures with hundreds of atoms and is the closest prior work to our unconditional all-atom MOF setting.

\paragraph{Point-Cloud Generative Models.}
PointFlow \citep{yangPointFlow3DPoint2019} combines a global shape latent with a point-level flow, while LION \citep{zengLIONLatentPoint2022} diffuses global and point-structured latents. Equivariant flow matching \citep{kleinEquivariantFlowMatching2023} aligns source particles and target points using OT. Not-So-Optimal Transport \citep{huiNotSoOptimalTransportFlows2025} studies how coupling redistributes vector-field complexity. Wasserstein Flow Matching \citep{havivWassersteinFlowMatching2025} treats point clouds as empirical distributions, and Orbit-Space Particle Flow Matching \citep{wangGenerativeModelingOrbitSpace2026} uses a representation of the permutation quotient.

Global-code point generation predates these methods: PSGN \citep{fanPointSetGeneration2017} predicts a point set from an image representation; \citet{achlioptasLearningRepresentationsGenerative2018} learn generative models over global point-cloud latents; FoldingNet \citep{yangFoldingNetPointCloud2018} decodes through fixed grid seeds; and PCGen \citep{verchevalPCGenFullyParallelizable2024} generates point clouds in parallel from compact global variables.

\paragraph{General Set Models.}
Deep Sets \citep{zaheerDeepSets2017} and Set Transformer \citep{leeSetTransformerFramework2019} provide invariant aggregation and attention-based set processing. Fixed-dimensional continuous representations are not universally lossless; worst-case latent-width requirements can grow with cardinality \citep{wagstaffLimitationsRepresentingFunctions2019,wagstaffUniversalApproximationFunctions2022}.

FSPool \citep{zhangFSPOOLLEARNINGSET2020} identifies the responsibility problem in set reconstruction. DSPN \citep{zhangDeepSetPrediction2019} reconstructs sets by optimization, whereas TSPN \citep{kosiorekConditionalSetGeneration2020} predicts them in one Transformer pass. PISA \citep{kortvelesyPermutationInvariantSetAutoencoders2023} combines a global embedding with deterministic queries; Top-N \citep{vignacTopNEquivariantSet2022} uses latent-conditioned reference elements. Learned queries and matching also appear in DETR \citep{carionEndtoEndObjectDetection2020}, and Slot Attention \citep{locatelloObjectcentricLearningSlot2020} learns object-centric slots. SetVAE \citep{kimSetVAELearningHierarchical2021} uses hierarchical set latents, while GraphVAE \citep{simonovskyGraphVAEGenerationSmall2018} decodes a global latent and aligns outputs by graph matching.

\paragraph{Permutation Ambiguity and Symmetry.}
Permutation-Symmetrized Diffusion \citep{koPermutationSymmetrizedDiffusionUnconditional2026}, PolyDiffuse \citep{chenPolyDiffusePolygonalShape2023}, and SwinGNN \citep{yanSwinGNNRethinkingPermutation2024} address equivalent indexed representations in molecular, set, and graph generation. Deterministic equivariant functions cannot distinguish elements with identical complete states \citep{kabaSymmetryBreakingEquivariant2024,zhangMultisetEquivariantSetPrediction2022}, and continuous canonicalization has limitations under common symmetry groups \citep{dymEquivariantFramesImpossibility2024}. GLASS builds on invariant representation and set decoding to combine global latent transport with parallel all-atom reconstruction, without requiring a canonical atom order.

\section{Additional theory for the correspondence problem}
\label{app:correspondence-theory}

\subsection{Assignment boundaries and exact OT paths}
\label{app:assignment-boundary}

For source configuration $\mathbf{x}_0$ and target set $\mathbf{y}$, OT selects $\pi^\star\in\arg\min_{\pi\in S_N}C_\pi(\mathbf{x}_0)$, where $C_\pi(\mathbf{x}_0)=\|\mathbf{x}_0-\mathbf{y}^\pi\|_F^2$. For two source particles at positions $x_1,x_2$ and target points $y_1,y_2$, expanding the identity and swapped costs gives
\begin{equation}
    C_{\mathrm{id}}-C_{\mathrm{swap}}
    =2(x_1-x_2)^\top(y_2-y_1).
    \label{eq:assignment_boundary}
\end{equation}
The assignment switches across $(x_1-x_2)^\top(y_1-y_2)=0$. For example, take $y_1=(-1,0)$, $y_2=(1,0)$ and $x_1=(\varepsilon,1)$, $x_2=(-\varepsilon,-1)$. The first particle is assigned to $y_2$ for $\varepsilon>0$ and to $y_1$ for $\varepsilon<0$, with initial velocity limits $(1,-1)$ and $(-1,-1)$. The discontinuity is in the dependence on source positions; each path remains continuous in time.

For a fixed target set with distinct sites, define $\mathbf{x}_t=(1-t)\mathbf{x}_0+t\mathbf{y}^{\pi^\star}$, with $0<t<1$. For any $\sigma\neq\pi^\star$,
\begin{equation}
\begin{aligned}
    C_\sigma(\mathbf{x}_t)-C_{\pi^\star}(\mathbf{x}_t)
    &=(1-t)\bigl[C_\sigma(\mathbf{x}_0)-C_{\pi^\star}(\mathbf{x}_0)\bigr]\\
    &\quad+t\|\mathbf{y}^{\pi^\star}-\mathbf{y}^\sigma\|_F^2>0.
\end{aligned}
\end{equation}
Optimality makes the first term nonnegative; distinct target sites make the second positive. Matching $\mathbf{x}_t$ to the target therefore uniquely recovers $\pi^\star$, then $\mathbf{x}_0=(\mathbf{x}_t-t\mathbf{y}^{\pi^\star})/(1-t)$ and the prescribed velocity.

For two particles assigned to distinct target sites, set $a=x_{0,i}-x_{0,j}$ and $b=y_{\pi^\star(i)}-y_{\pi^\star(j)}$. Optimality against swapping their destinations gives $a^\top b\geq0$, so
\begin{equation}
    \|x_{t,i}-x_{t,j}\|_2^2
    =(1-t)^2\|a\|_2^2+2t(1-t)a^\top b+t^2\|b\|_2^2
    \geq t^2\|b\|_2^2.
\end{equation}
Thus, particles following exact OT paths remain separated for $t>0$.

\subsection{Equivariance and local sensitivity}
\label{app:equivariance-sensitivity}

We specialize the symmetry constraints of \citet{zhangMultisetEquivariantSetPrediction2022,kabaSymmetryBreakingEquivariant2024}. Let $x_i$ denote the complete state of particle $i$, with time and shared conditioning fixed, and let $f$ be deterministic and permutation equivariant. If the permutation $P_{ij}$ exchanging particles $i,j$ satisfies $P_{ij}\mathbf{x}=\mathbf{x}$, then $f(\mathbf{x})=P_{ij}f(\mathbf{x})$, giving $f_i(\mathbf{x})=f_j(\mathbf{x})$.

If $f$ is also $L$-Lipschitz in the Frobenius norm on a permutation-invariant domain, then
\begin{equation}
    \|f(\mathbf{x})-P_{ij}f(\mathbf{x})\|_F
    \leq L\|\mathbf{x}-P_{ij}\mathbf{x}\|_F,
\end{equation}
which gives $\|f_i(\mathbf{x})-f_j(\mathbf{x})\|_2\leq L\|x_i-x_j\|_2$. Now let $f$ predict endpoint positions, let $y_i,y_j$ denote its assigned target sites separated by $\Delta$, and suppose each endpoint error is at most $\eta$. The triangle inequality yields
\begin{equation}
\begin{aligned}
    \Delta
    &\leq\|y_i-f_i(\mathbf{x})\|_2+\|f_i(\mathbf{x})-f_j(\mathbf{x})\|_2
      +\|f_j(\mathbf{x})-y_j\|_2\\
    &\leq2\eta+L\delta,\qquad \delta=\|x_i-x_j\|_2.
\end{aligned}
\end{equation}
For an $L_v$-Lipschitz equivariant velocity predictor, the same triangle-inequality argument applies to prescribed velocities $u_i,u_j$. If each velocity error is at most $\eta_v$, then
\begin{equation}
    \Delta_v\leq2\eta_v+L_v\delta,
    \qquad \Delta_v=\|u_i-u_j\|_2.
\end{equation}
Both bounds require sufficient predictor sensitivity when nearby states have well-separated targets. They constrain prediction accuracy, not training time. If the endpoint predictor is the complete sampling map, $L$ is the Lipschitz constant of that map.

\section{Additional details for the fixed-grid experiment}
\label{app:toy-experiments}

\subsection{Target grids and training protocol}
\label{app:toy-grid}

We use $N\in\{4,8,12,16,20,24,28,32,40,48,64\}$ and three training seeds per coupling. All target grids lie in $[-1,1]^2$. For each $N$, we construct an $r\times c$ grid spanning this domain, with
\begin{equation}
\begin{aligned}
    r&=\lceil\sqrt{N}\rceil, & c&=\left\lceil\frac{N}{r}\right\rceil,\\
    \Delta x&=\frac{2}{c-1}, & \Delta y&=\frac{2}{r-1}.
\end{aligned}
\end{equation}
If $rc>N$, the surplus sites nearest the origin are removed. The displayed cases use a full $2\times2$ grid at $N=4$, a $5\times5$ grid with its center removed at $N=24$, and a full $8\times8$ grid at $N=64$. The domain area remains $4$, the site density is $N/4$, and minimum target spacing decreases approximately as $2/\sqrt{N}$.

Each flow uses a three-layer permutation-equivariant Transformer with width 64, four attention heads, feedforward width 128, GELU activations, pre-layer normalization, and no dropout or particle-index embeddings. The input concatenates each particle's two coordinates with time; a linear output head predicts its two velocity components. We train in FP32 with AdamW, learning rate $10^{-3}$, weight decay $10^{-6}$, batch size 512, and gradient norm clipping at 1.

Independent coupling draws a fresh random target permutation for each training pair. OT coupling uses the Hungarian algorithm to minimize the sum of squared source-to-target distances within each configuration. Training times are uniform on $[0,1)$, and the velocity loss is averaged over samples, particles, and coordinates.

The flow runs in Figure~\ref{fig:toy} stop at the first evaluation with mean absolute permutation-invariant RMSD (PI-RMSD) at most $0.01$, or at 1M updates. For one generated set,
\begin{equation}
    \operatorname{PI\text{-}RMSD}(\hat{\mathbf{x}},\mathbf{y})
    =\left[\frac{1}{N}\min_{\pi\in S_N}
    \sum_i\|\hat{x}_i-y_{\pi(i)}\|_2^2\right]^{1/2}.
\end{equation}
This stopping tolerance is in the coordinates of $[-1,1]^2$ and is not divided by grid spacing. Evaluation uses 64 uniform Euler steps and a fixed set of $1,024$ Gaussian source configurations per run, separate from the training pool. Evaluations occur every 100 updates below 20k, every 500 below 100k, every 2k below 500k, and every 5k thereafter.

\subsection{Slot-decoder control}
\label{app:toy-decoder}

\begin{figure}[t]
    \centering
    \includegraphics[width=\linewidth]{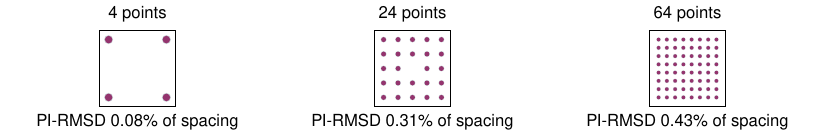}
    \caption{\textbf{Slot-decoder outputs on the fixed target grids.} For each displayed $N$, the run with the largest final PI-RMSD among three seeds, stated relative to the minimum target-site spacing. No target site is shared.}
    \label{fig:toy-decoder}
\end{figure}

The slot decoder is the GLASS decoder without a global latent. Each of $N$ learned 64-dimensional slot embeddings is linearly projected to width 64 and processed by three pre-layer-norm self-attention layers with four heads and SwiGLU feedforward blocks, followed by a final layer norm and a linear head predicting two coordinates. It is trained on the same fixed target grids with the flows' optimizer settings, one target set per step, and a mean-squared error to the target sites. At every step, the target sites are randomly permuted and matched to the predicted positions with the Hungarian algorithm, as in autoencoder training. The decoder output is deterministic, so each evaluation scores one set with the metrics below. Training stops at the first evaluation, every 100 steps, at which the mean PI-RMSD is at most $0.01$ of the minimum target-site spacing. For all $N$ and seeds, this occurs at the first evaluation, where both failure thresholds are also met, so the reported training steps do not depend on the stopping rule. Figure~\ref{fig:toy-decoder} shows the worst of the three runs at each displayed $N$.

\subsection{Occupancy metrics and training-step summaries}
\label{app:site-coverage}

For $N$ generated particles and $N$ target sites, define the nearest-site assignment $a(i)=\arg\min_j\|\hat{x}_i-y_j\|_2$, choosing the lowest target index in a tie, and
\begin{equation}
    m(\hat{\mathbf{x}},\mathbf{y})
    =1-\frac{|\{a(i):i=1,\ldots,N\}|}{N}.
    \label{eq:unoccupied-sites}
\end{equation}
Here $m$ is the unoccupied-site fraction, and a set fails when $m>0$. These metrics use no distance threshold: distinct generated particles can share a nearest site, and unique occupancy does not bound positional error.

For each run, we record the first evaluated training step at which the failure rate is below $50\,\%$ or $10\,\%$. Figure~\ref{fig:toy}b reports the mean and standard deviation across three runs, assigning unreached thresholds the 1M-step budget and marking them with hollow triangles.

At each flow run's final checkpoint, we average $m$ over all $1,024$ generated sets, including those with $m=0$, and report the mean and standard deviation across runs. Final checkpoints follow the stopping rule above and can differ in training duration. For the flows, the mean unoccupied-site fraction rises from near zero at small $N$ to approximately $2$--$3\,\%$ at $N=64$; the slot decoder leaves no site unoccupied at any $N$.

\subsection{Sampling-resolution control}
\label{app:sampling-resolution}

We evaluate retained checkpoints at $N=4,24,64$ from one training seed with multiple Euler resolutions, using the same initial samples for each checkpoint. The metric is the percentage of generated sets with duplicate-site occupancy.

\begin{table}[t]
    \centering
    \caption{Failure rates (\%) at different sampling resolutions for one training run per setting. Dashes denote resolutions not evaluated.}
    \label{tab:sampling-resolution}
    \small
    \begin{tabular}{llrrr}
        \toprule
        $N$ & Coupling & 64 steps & 1024 steps & 4096 steps \\
        \midrule
        4  & Independent & 0.00 & 0.00 & -- \\
        4  & OT          & 0.00 & 0.00 & -- \\
        24 & Independent & 10.16 & 5.86 & 5.86 \\
        24 & OT          & 20.31 & 19.92 & 19.92 \\
        64 & Independent & 80.08 & 58.20 & 57.03 \\
        64 & OT          & 88.28 & 87.50 & 87.50 \\
        \bottomrule
    \end{tabular}
\end{table}

Finer integration improves the larger independent-coupling models, while OT changes little. For both couplings, the change from 1024 to 4096 steps is small and the cardinality dependence remains.

\FloatBarrier
\section{Experimental Details and Model Settings}
\label{sec:experimental_details}
\label{sec:model_settings}

\FloatBarrier
\subsection{Datasets and Representation}
\label{app:glass-details}

MP20 uses the CDVAE split of $27,136/9,047/9,046$ training/validation/test structures. QMOF150 uses the ADiT split of $14,589/1,024/1,024$ structures with at most 150 atoms per cell. We use periodic fractional coordinates without translation augmentation or origin alignment. The encoder masks padding and pools over real atoms; the decoder predicts atomic numbers $1,\ldots,94$ and empty class $0$. Fractional coordinates are wrapped into $[0,1)$.

\FloatBarrier
\subsection{Training and Sampling}
\label{app:mp20-checkpoints}

We train the autoencoder, freeze it, and standardize its latents using training-set means and standard deviations before fitting the flow. Tables~\ref{tab:ae_settings} and~\ref{tab:flow_settings} give the settings. Reconstruction losses pool their sums and normalization counts across the minibatch.

We use the 50k flow checkpoint for MP20 to balance validity and novelty, including all MP20 ablations, and the 1M checkpoint for QMOF150. Main comparisons average three independent training runs; ablations use one run per setting. Reported $\pm$ values are sample standard deviations across runs.

\begin{table}[H]
\centering
\caption{\textbf{Autoencoder settings.} Ablations vary the indicated component.}
\label{tab:ae_settings}
\begingroup
\small
\setlength{\tabcolsep}{4pt}
\renewcommand{\arraystretch}{1}
\glassfit{%
\begin{tabular}{lcc}
\toprule
Setting & MP20 & QMOF150 \\
\midrule
Model dimension & 256 & 256 \\
Latent dimension & 16 & 64 \\
Encoder layers & 4 & 4 \\
Decoder layers & 6 & 12 \\
Attention heads & 8 & 8 \\
Decoder slot dimension & 32 & 32 \\
Coordinate Fourier bands & 4 & 4 \\
Training steps & 100k & 100k \\
Batch size & 256 & 256 \\
Peak learning rate & $3\times10^{-4}$ & $3\times10^{-4}$ \\
Warmup steps & 2k & 1k \\
Optimizer & AdamW & AdamW \\
Weight decay & 0 & 0 \\
Training precision & FP32 & FP32 \\
Coordinate / type / lattice / pair loss weights
    & 4.0 / 0.05 / 5.0 / 0.5
    & 1.0 / 0.05 / 5.0 / 0.5 \\
\bottomrule
\end{tabular}%
}
\endgroup
\end{table}

\begin{table}[H]
\centering
\caption{\textbf{Latent flow settings.} Sampling steps count midpoint integration steps.}
\label{tab:flow_settings}
\begingroup
\small
\setlength{\tabcolsep}{4pt}
\renewcommand{\arraystretch}{1}
\glassfit{%
\begin{tabular}{lcc}
\toprule
Setting & MP20 & QMOF150 \\
\midrule
Hidden dimension & 768 & 768 \\
Residual blocks & 8 & 8 \\
Training steps & 1M & 1M \\
Batch size & 1024 & 1024 \\
Peak learning rate & $3\times10^{-4}$ & $3\times10^{-4}$ \\
Warmup steps & 2k & 2k \\
Optimizer & AdamW & AdamW \\
Weight decay & 0 & 0 \\
EMA decay & 0.9999 & 0.9999 \\
Training precision & BF16 & BF16 \\
Sampling steps & 32 & \QmofSamplingSteps{} \\
Integrator & Midpoint & Midpoint \\
Integration / decoding precision & FP32 & FP32 \\
\bottomrule
\end{tabular}%
}
\endgroup
\end{table}

\FloatBarrier
\subsection{Reconstruction Metrics}
\label{app:cell-reconstruction}

Reconstruction RMSD uses the predicted fractional coordinates and predicted lattice, with the reconstruction objective's atom assignment. We average per-structure Cartesian RMSDs over the split, using a common lower-triangular cell orientation and the objective's 27-image periodic search, without rescaling or origin alignment. Type accuracy is pooled over real target atoms; padding is excluded from both metrics.

\FloatBarrier
\subsection{Geometry Relaxation}
\label{sec:relaxation_protocol}

MP20 pre-relaxation uses NequIP-OAM-L \citep{batznerE3equivariantGraphNeural2022,kavanaghNequIPAllegroFoundation2026} with ASE FIRE \citep{hjorthlarsenAtomicSimulationEnvironment2017} and a Fréchet cell filter, at most 200 steps, and $f_{\max}=0.005$\,eV/\AA. LeMat's internal relaxation is specified separately below.

GLASS-opt on QMOF150 uses eSEN-30M-OAM \citep{fuLearningSmoothExpressive2025} plus two-body D3(BJ) \citep{grimmeConsistentAccurateInitio2010,grimmeEffectDampingFunction2011}, with PBE parameters and 40-Bohr dispersion and coordination cutoffs. We selected eSEN-OAM+D3 based on its performance in MOFSimBench \citep{krassMOFSimBenchEvaluatingUniversal2025a}. Both terms contribute energy, forces, and stress. Batched TorchSim \citep{cohenTorchSimEfficientAtomistic2025} L-BFGS optimizes atoms and the cell through a Frechet filter, with at most 200 steps, maximum step 0.1, and $f_{\max}=0.02$\,eV/\AA. Relaxation displacement RMSD uses fixed atom correspondence and final-cell periodic images, including cell deformation. Failed candidates remain in requested-sample yield denominators.

\FloatBarrier
\Needspace{12\baselineskip}
\subsection{LeMat-GenBench Evaluation}
\label{app:lemat-protocol}

LeMat-GenBench evaluation \citep{betalaLeMatGenBenchUnifiedEvaluation2025} uses the \texttt{comprehensive\_multi\_mlip\_hull} protocol and the complete LeMat-Bulk \texttt{compatible\_pbe} novelty reference. Validity checks use charge tolerance $0.1$, distance scaling $0.5$, atomic-density bounds $[10^{-5},0.5]$\,atoms/\AA$^3$, mass-density bounds $[0.01,25]$\,g/cm$^3$, and format and symmetry checks. Novelty and uniqueness use StructureMatcher fingerprints with tolerance $0.1$.

Energy scoring uses Orb-v3-conservative-inf-OMat \citep{rhodesOrbv3AtomisticSimulation2025}, MACE-MP-0b3 \citep{batatiaMACEHigherOrder2023,batatiaFoundationModelAtomistic2024}, and UMA-s1p1 \citep{woodUMAFamilyUniversal2025} with the OMat task and their respective reference hulls. The ensemble requires two usable models, with the evaluator's single-model fallback when necessary. Hull energies refer to the supplied geometry; internal relaxation uses $f_{\max}=0.02$\,eV/\AA\ and at most 50 steps to measure displacement. SUN requires $E_{\rm hull}\leq0$, while MSUN uses the disjoint range $0<E_{\rm hull}\leq0.1$\,eV/atom.

LeMat evaluates novelty among valid candidates. We distinguish conditional novelty $N\mid V$ from valid-and-novel yield $|V\cap N|/n$, where $n$ is the requested sample count. Joint yields retain failed inputs; mean energies and relaxation displacement use successfully scored structures. Checkpoint sweeps and ablations report LeMat metrics without internal relaxation or its displacement metric.

\FloatBarrier
\subsection{Crystalite Evaluation}
\label{sec:crystalite_evaluation}

The Crystalite DNG protocol \citep{veljkovicCrystaliteLightweightTransformer2026} defines structural validity by cell volume $\geq0.1$\,\AA$^3$ and minimum interatomic distance $\geq0.5$\,\AA; composition validity uses SMACT \citep{daviesSMACTSemiconductingMaterials2019}. StructureMatcher uses site tolerance 0.5, lattice tolerance 0.3, and angle tolerance $10^\circ$. Novelty uses the official Crystalite training reference ($27,138$ structures), and density and element-count Wasserstein distances use its validation reference ($9,046$ structures). Metrics use the evaluator's corresponding constructed and valid pools.

Thermodynamic scoring uses NequIP-OAM-L, the MP2020-like hull procedure, and the 200-step FIRE/Frechet relaxation above. Crystalite calls $E_{\rm hull}\leq0.1$\,eV/atom ``Stable'' and its stable, unique, novel intersection ``S.U.N.''; this differs from LeMat's strict SUN definition. Results are shown in section~\ref{app:crystalite_evaluation_results}.

\FloatBarrier
\section{Architecture Ablations}
\label{sec:architecture-ablations}

\FloatBarrier
\subsection{MP20 Autoencoder Capacity}

We vary latent dimension, encoder depth, and decoder depth around the reference configuration ($d_z=16$, four encoder layers, six decoder layers), reporting full-split reconstruction and LeMat metrics on $2,500$ generated structures per setting. Larger latents and deeper decoders improve reconstruction, but generation also shifts along the validity--novelty tradeoff.

\begin{table}[H]
\centering
\caption{\textbf{Latent dimension.} Encoder and decoder depths are fixed to 4 and 6.}
\label{tab:latent-dimension-ablation}
\label{tab:ablation_latent}
\begingroup
\small
\setlength{\tabcolsep}{4pt}
\renewcommand{\arraystretch}{1}
\glassfit{%
\begin{tabular}{lrrrrrrr}
\toprule
& \multicolumn{2}{c}{AE training split} & \multicolumn{2}{c}{AE validation split} & \multicolumn{3}{c}{Generation (50k)} \\
\cmidrule(lr){2-3}\cmidrule(lr){4-5}\cmidrule(lr){6-8}
Latent dim. & RMSD (\AA) & Type acc. (\%) & RMSD (\AA) & Type acc. (\%) & Valid (\%) & Novel $\mid$ valid (\%) & Valid $\cap$ novel (\%) \\
\midrule
8 & 0.1064 & 100.00 & 0.3547 & 55.60 & 78.60 & 62.54 & 49.16 \\
16 & 0.0339 & 100.00 & 0.3001 & 63.26 & 88.08 & 51.09 & 45.00 \\
32 & 0.0159 & 100.00 & 0.2696 & 72.36 & 91.92 & 38.29 & 35.20 \\
\bottomrule
\end{tabular}
}
\endgroup
\end{table}

\begin{table}[H]
\centering
\caption{\textbf{Encoder depth.} Latent dimension and decoder depth are fixed to 16 and 6.}
\label{tab:encoder-depth-ablation}
\label{tab:ablation_encoder}
\begingroup
\small
\setlength{\tabcolsep}{4pt}
\renewcommand{\arraystretch}{1}
\glassfit{%
\begin{tabular}{lrrrrrrr}
\toprule
& \multicolumn{2}{c}{AE training split} & \multicolumn{2}{c}{AE validation split} & \multicolumn{3}{c}{Generation (50k)} \\
\cmidrule(lr){2-3}\cmidrule(lr){4-5}\cmidrule(lr){6-8}
Encoder layers & RMSD (\AA) & Type acc. (\%) & RMSD (\AA) & Type acc. (\%) & Valid (\%) & Novel $\mid$ valid (\%) & Valid $\cap$ novel (\%) \\
\midrule
2 & 0.0386 & 100.00 & 0.2837 & 63.33 & 89.24 & 50.34 & 44.92 \\
4 & 0.0339 & 100.00 & 0.3001 & 63.26 & 88.08 & 51.09 & 45.00 \\
6 & 0.0355 & 100.00 & 0.2980 & 63.25 & 89.04 & 55.97 & 49.84 \\
\bottomrule
\end{tabular}
}
\endgroup
\end{table}

\begin{table}[H]
\centering
\caption{\textbf{Decoder depth.} Latent dimension and encoder depth are fixed to 16 and 4.}
\label{tab:decoder-depth-ablation}
\label{tab:ablation_decoder}
\begingroup
\small
\setlength{\tabcolsep}{4pt}
\renewcommand{\arraystretch}{1}
\glassfit{%
\begin{tabular}{lrrrrrrr}
\toprule
& \multicolumn{2}{c}{AE training split} & \multicolumn{2}{c}{AE validation split} & \multicolumn{3}{c}{Generation (50k)} \\
\cmidrule(lr){2-3}\cmidrule(lr){4-5}\cmidrule(lr){6-8}
Decoder layers & RMSD (\AA) & Type acc. (\%) & RMSD (\AA) & Type acc. (\%) & Valid (\%) & Novel $\mid$ valid (\%) & Valid $\cap$ novel (\%) \\
\midrule
3 & 0.0622 & 100.00 & 0.3420 & 60.02 & 87.80 & 58.63 & 51.48 \\
6 & 0.0339 & 100.00 & 0.3001 & 63.26 & 88.08 & 51.09 & 45.00 \\
9 & 0.0271 & 100.00 & 0.2679 & 65.14 & 90.48 & 48.98 & 44.32 \\
\bottomrule
\end{tabular}
}
\endgroup
\end{table}

\FloatBarrier
\Needspace{0.45\textheight}
\subsection{Variational Bottleneck on MP20}
\label{app:mp20-vae}

With the same backbone and $d_z=16$, the variational encoder predicts a diagonal-Gaussian posterior and decodes a reparameterized sample during training:
\begin{equation}
\mathcal L_{\mathrm{VAE}}=\mathcal L_{\mathrm{rec}}+\beta\,D_{\mathrm{KL}}\!\left(q(\mathbf z\mid X)\,\|\,\mathcal N(\mathbf 0,\mathbf I)\right),
\qquad \beta\in\{10^{-5},10^{-4}\}.
\end{equation}
The KL term is summed over latent dimensions and averaged over structures. Reconstruction evaluation and flow training use posterior means; generation uses the learned flow. This tests KL regularization of the representation. The weaker regularization slightly improves valid-and-novel yield, while the stronger setting increases validity at the expense of novelty (Table~\ref{tab:mp20-vae}).

\begin{table}[H]
\centering
\caption{\textbf{Variational bottleneck.} Complete-split reconstruction and LeMat metrics on $2,500$ generated structures per setting at 50k flow steps.}
\label{tab:mp20-vae}
\begingroup
\small
\setlength{\tabcolsep}{4pt}
\glassfit{%
\begin{tabular}{lrrrrrrr}
\toprule
& \multicolumn{2}{c}{Training split} & \multicolumn{2}{c}{Validation split} & \multicolumn{3}{c}{Generation} \\
\cmidrule(lr){2-3}\cmidrule(lr){4-5}\cmidrule(lr){6-8}
Model ($\beta$) & RMSD (\AA) & Type acc. (\%) & RMSD (\AA) & Type acc. (\%) & Valid (\%) & Novel $\mid$ valid (\%) & Valid $\cap$ novel (\%) \\
\midrule
Deterministic & 0.0339 & 100.00 & 0.3001 & 63.26 & 88.08 & 51.09 & 45.00 \\
VAE ($10^{-5}$) & 0.0415 & 100.00 & 0.2915 & 63.55 & 89.84 & 51.47 & 46.24 \\
VAE ($10^{-4}$) & 0.0548 & 100.00 & 0.2914 & 64.42 & 91.60 & 45.24 & 41.44 \\
\bottomrule
\end{tabular}
}
\endgroup
\end{table}

\FloatBarrier
\subsection{QMOF150 Autoencoder Capacity}
\label{sec:qmof_capacity}

Increasing the latent dimension from 32 to 128 leaves the large train--validation reconstruction gap largely unchanged (Table~\ref{tab:qmof150-latent-ae}).

\begin{table}[H]
\centering
\caption{\textbf{QMOF150 latent-dimension ablation.} Complete-split reconstruction after 100k autoencoder steps.}
\label{tab:qmof150-latent-ae}
\begingroup
\small
\setlength{\tabcolsep}{4pt}
\renewcommand{\arraystretch}{1}
\glassfit{%
\begin{tabular}{rrrrr}
\toprule
& \multicolumn{2}{c}{Training split} & \multicolumn{2}{c}{Validation split} \\
\cmidrule(lr){2-3}\cmidrule(lr){4-5}
Latent dim. & RMSD (\AA) & Type acc. (\%) & RMSD (\AA) & Type acc. (\%) \\
\midrule
32 & 0.0243 & 99.9990 & 1.2793 & 58.8782 \\
64 & 0.0190 & \textbf{99.9999} & \textbf{1.2473} & 58.6839 \\
128 & \textbf{0.0179} & \textbf{99.9999} & 1.2594 & \textbf{59.1796} \\
\bottomrule
\end{tabular}
}
\endgroup
\end{table}

\FloatBarrier
\subsection{LeMat-Bulk Autoencoders}
\label{app:lemat-bulk}

We train autoencoders on the LeMat-Bulk \texttt{compatible\_pbe} subset \citep{betalaLeMatGenBenchUnifiedEvaluation2025}, restricted to at most 20 atoms per structure. The split contains \LematBulkTrainStructures{} training structures and $10,000$ validation structures, with no shared BAWL fingerprints between the training and validation sets. Training uses 1M steps, batch size 256, learning rate $10^{-4}$, and MP20 loss weights with type weight $0.5$. All models have width 384, six encoder layers, eight decoder layers, and slot dimension 256.

The train--validation gap is much smaller than on MP20, and larger latents reduce reconstruction error (Table~\ref{tab:lemat-bulk-ae}). This larger, more diverse dataset also uses different training settings, so the comparison does not isolate dataset size.

\begin{table}[H]
\centering
\caption{\textbf{LeMat-Bulk autoencoders.} Reconstruction over the training (\LematBulkTrainStructures{} structures) and validation ($10,000$ structures) splits. Invalid predicted cells, at most $0.05\,\%$ of either split, are excluded from RMSD.}
\label{tab:lemat-bulk-ae}
\small
\begin{tabular}{rrrrr}
\toprule
& \multicolumn{2}{c}{Training split} & \multicolumn{2}{c}{Validation split} \\
\cmidrule(lr){2-3}\cmidrule(lr){4-5}
Latent dim. & RMSD (\AA) & Type acc. (\%) & RMSD (\AA) & Type acc. (\%) \\
\midrule
32 & 0.0905 & 99.92 & 0.1098 & 97.05 \\
64 & 0.0584 & 99.72 & 0.0637 & 98.85 \\
128 & 0.0434 & 99.75 & 0.0460 & 99.49 \\
\bottomrule
\end{tabular}
\end{table}

\FloatBarrier
\Needspace{0.55\textheight}
\section{Additional MP20 Results}
\label{sec:mp20_checkpoint_analysis}

\FloatBarrier
\subsection{Validity--Novelty Tradeoff}
\label{sec:glass_validity_novelty_tradeoff}

\begin{figure}[H]
\centering
\includegraphics[width=.72\linewidth]{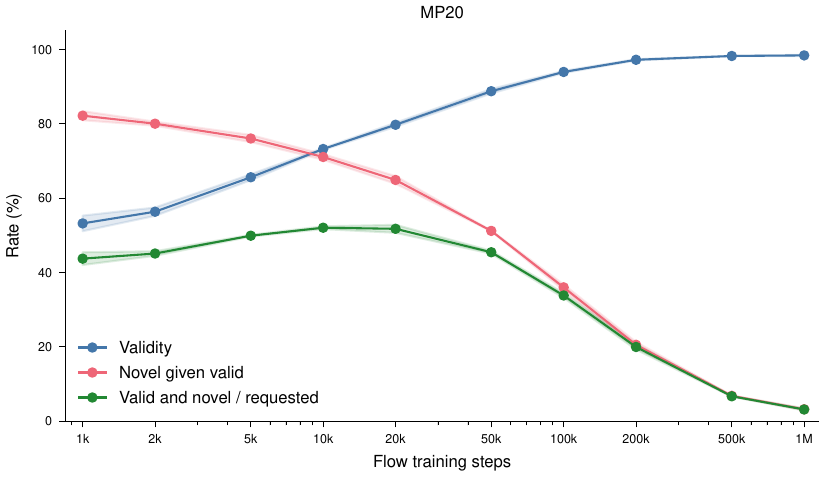}
\caption{\textbf{MP20 validity--novelty tradeoff.} Raw validity, novelty among valid candidates, and their joint yield across flow checkpoints. Curves show means and sample standard deviations over three training runs, with $2,500$ samples per run and checkpoint.}
\label{fig:checkpoint_tradeoff}
\end{figure}

Longer flow training increases validity while reducing novelty (Figure~\ref{fig:checkpoint_tradeoff}). The 50k checkpoint balances the two; the 1M checkpoint approaches training-set validity with much lower novel yield.

\FloatBarrier
\Needspace{0.7\textheight}
\subsection{Extended LeMat-GenBench Comparison}

Table~\ref{tab:lemat-genbench-full} extends the main comparison to all reference methods, using the protocol in Appendix~\ref{app:lemat-protocol}. Additional models include MCFlow \citep{seongMultimodalCrystalFlow2026}, PLaID++ \citep{xuPLaIDPreferenceAligned2026}, WyFormer \citep{kazeevWyckoffTransformerGeneration2025}, Chemeleon1 \citep{parkExplorationCrystalChemical2025}, DiffCSP++ \citep{jiaoSpaceGroupConstrained2023}, CrystaLLM-pi \citep{boneDiscoveryRecoveryCrystalline2026}, and CrystalFormer \citep{caoSpaceGroupInformed2025}.

\begin{table}[H]
\centering
\caption{\textbf{Extended LeMat-GenBench comparison.} GLASS reports means and sample standard deviations over three runs at 50k, with $2,500$ samples per run. Reference results are from the benchmark leaderboard \citep{betalaLeMatGenBenchUnifiedEvaluation2025}. Groups distinguish raw and NequIP-OAM-L-pre-relaxed inputs. MSUN excludes SUN; bold marks the best mean within each group.}
\label{tab:lemat-genbench-full}
\begingroup
\small
\setlength{\tabcolsep}{4pt}
\renewcommand{\arraystretch}{1}
\glassfit{%
\begin{tabular}{lrrrrrrrrr}
\toprule
Model & Valid & Unique & Novel & Stable & Metastable & SUN & MSUN & E Above Hull & Relax. RMSD \\
& (\%) $\uparrow$ & (\%) $\uparrow$ & (\%) $\uparrow$ & (\%) $\uparrow$ & (\%) $\uparrow$ & (\%) $\uparrow$ & (\%) $\uparrow$ & (eV/atom) $\downarrow$ & (\AA) $\downarrow$ \\
\midrule
\multicolumn{10}{c}{\textbf{Pre-relaxed inputs}} \\
\midrule
Crystalite & \textbf{97.20} & 95.80 & 53.20 & 12.70 & 51.60 & 1.50 & \textbf{22.60} & 0.0905 & 0.1322 \\
MCFlow & \textbf{97.20} & \textbf{96.30} & 52.20 & 11.90 & 49.30 & 0.70 & 18.90 & 0.0987 & 0.1696 \\
OMatG & 96.40 & 95.20 & 51.20 & 11.60 & 49.80 & 1.00 & 18.00 & 0.0956 & 0.0759 \\
MiAD & 96.20 & 94.30 & 40.20 & 6.40 & 62.00 & 1.00 & 16.60 & 0.0804 & 0.2494 \\
MatterGen & 95.70 & 95.10 & \textbf{70.50} & 2.00 & 33.40 & 0.20 & 15.00 & 0.1834 & 0.3878 \\
OMatG-FC & \textbf{97.20} & 92.80 & 28.90 & \textbf{18.40} & 57.90 & 1.70 & 12.00 & \textbf{0.0694} & \textbf{0.0685} \\
PLaID++ & 96.00 & 77.80 & 24.20 & 12.40 & 60.70 & 1.00 & 7.60 & 0.0854 & 0.1286 \\
WyFormer & 93.40 & 93.00 & 66.40 & 0.50 & 15.70 & 0.10 & 1.90 & 0.4988 & 0.8121 \\
\midrule
GLASS + NequIP & 96.56 & 95.48 & 52.71 & 11.75 & \textbf{63.08} & \textbf{1.92} & 21.16 & 0.1261 & 0.1258 \\
 & $\pm$ 0.39 & $\pm$ 0.66 & $\pm$ 1.40 & $\pm$ 0.51 & $\pm$ 1.52 & $\pm$ 0.18 & $\pm$ 1.40 & $\pm$ 0.0080 & $\pm$ 0.0071 \\
\midrule
\multicolumn{10}{c}{\textbf{Inputs without pre-relaxation}} \\
\midrule
Chemeleon2 & 95.20 & 88.10 & \textbf{71.60} & 0.00 & 39.80 & 0.00 & \textbf{21.20} & \textbf{0.1557} & 0.4226 \\
Chemeleon1 & 95.40 & 94.80 & 64.00 & \textbf{3.60} & 34.40 & \textbf{0.30} & 11.30 & 0.2055 & 0.4907 \\
DiffCSP & \textbf{95.70} & 94.80 & 66.20 & 2.30 & 29.80 & 0.10 & 8.50 & 0.2747 & 0.5857 \\
DiffCSP++ & 95.30 & \textbf{95.10} & 62.00 & 1.00 & 26.40 & 0.20 & 5.00 & 0.4093 & 0.6933 \\
CrystaLLM-pi & 86.80 & 84.90 & 24.90 & 2.90 & \textbf{49.00} & \textbf{0.30} & 3.60 & 0.3205 & \textbf{0.3631} \\
CrystalFormer & 69.90 & 69.40 & 31.80 & 1.40 & 28.80 & 0.00 & 3.10 & 0.7039 & 0.6585 \\
ADiT & 90.60 & 87.80 & 26.00 & 0.40 & 36.50 & 0.00 & 1.00 & 0.3333 & 0.3794 \\
\midrule
GLASS & 89.29 & 88.32 & 45.65 & 0.93 & 41.95 & 0.15 & 11.49 & 0.3140 & 0.4290 \\
 & $\pm$ 0.59 & $\pm$ 0.68 & $\pm$ 1.20 & $\pm$ 0.10 & $\pm$ 1.65 & $\pm$ 0.06 & $\pm$ 0.61 & $\pm$ 0.0083 & $\pm$ 0.0077 \\
\bottomrule
\end{tabular}
}%
\endgroup
\end{table}

\FloatBarrier
\Needspace{0.55\textheight}
\subsection{Size-Resolved Validity and Local Geometry}
\label{app:mp20_validity_full}

\begin{figure}[H]
    \centering
    \includegraphics[width=\linewidth]{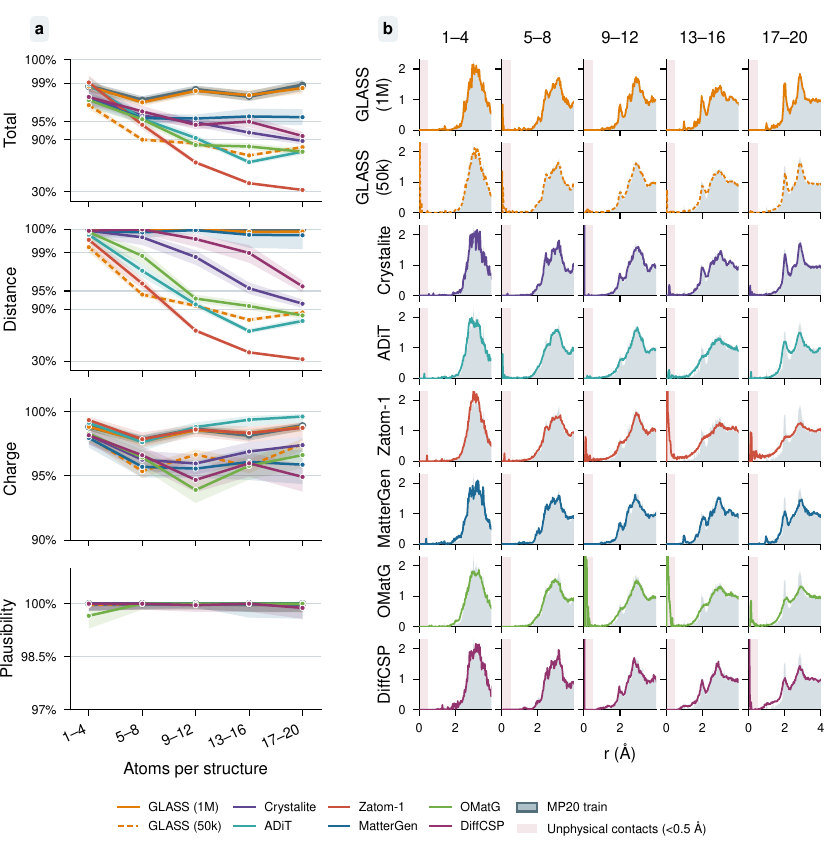}
    \caption{\textbf{MP20 validity and radial distribution functions by size.} RDFs show interatomic pair-distance distributions for generated structures and the training reference.}
    \label{fig:mp20_comparison_genbench}
\end{figure}

Table~\ref{tab:mp20-lemat-validity} separates the validity checks by atom count. Figure~\ref{fig:mp20_comparison_genbench} compares pair-distance distributions with all $27,136$ MP20 training structures. Both use $10,000$ GLASS samples per training run, pooled over three runs. Samples for MatterGen and DiffCSP were obtained from the released CrystalDiT \citep{yiCrystalDiTDiffusionTransformer2026} repository. Crystalite and OMatG were sampled using the released checkpoint, and ADiT and Zatom-1 were obtained from the respective repositories.

The Crystalite curves use 10,000 samples from the released MP20 checkpoint with its production sampler: EMA weights, 150 Heun steps on the Karras schedule, churn, and coordinate/lattice anti-annealing (Appendix~\ref{app:crystalite-churn}). For the OMatG curves, we generated 10,000 raw structures with the released MP-20-DNG Linear-SDE-Gamma checkpoint and configuration using the official \texttt{omg predict} sampler (710 integration steps). Its atom-count inputs comprise the 9,046 bundled MP20 test counts plus 954 resampled counts.

\begin{table}[H]
\centering
\caption{LeMat-GenBench validity rates (\%) corresponding exactly to the curves in the MP-20 validity plot. Overall rates include every requested structure; atom-count bins use inclusive bounds. The GT row is the plot's 27,136-structure MP-20 training split. Bold values are the best generated-model result in each column.}
\label{tab:mp20-lemat-validity}
\begin{minipage}[t]{0.49\textwidth}
\centering
\textbf{Total validity}\\[2pt]
\resizebox{\linewidth}{!}{%
\begin{tabular}{lrrrrrr}
\toprule
Model & Overall & 1--4 & 5--8 & 9--12 & 13--16 & 17--20 \\
\midrule
MP-20 train (GT) & 98.45 & 98.83 & 97.83 & 98.63 & 98.18 & 98.90 \\
\midrule
\textbf{GLASS (1M)} & \textbf{98.38} & 98.83 & \textbf{97.64} & \textbf{98.55} & \textbf{98.25} & \textbf{98.73} \\
\textbf{GLASS (50k)} & 89.23 & 97.36 & 90.07 & 88.52 & 82.00 & 86.79 \\
Crystalite & 94.44 & 98.11 & 96.00 & 94.86 & 92.40 & 89.54 \\
ADiT & 90.24 & 98.94 & 95.49 & 90.65 & 76.79 & 84.39 \\
Zatom-1 & 73.52 & \textbf{99.04} & 94.33 & 76.51 & 48.89 & 34.17 \\
MatterGen & 96.05 & 97.91 & 95.60 & 95.54 & 95.88 & 95.76 \\
OMatG & 90.96 & 97.85 & 95.41 & 87.82 & 87.12 & 84.33 \\
DiffCSP & 95.17 & 98.12 & 96.61 & 94.32 & 94.98 & 91.33 \\
\bottomrule
\end{tabular}%
}
\end{minipage}
\hfill
\begin{minipage}[t]{0.49\textwidth}
\centering
\textbf{Distance validity}\\[2pt]
\resizebox{\linewidth}{!}{%
\begin{tabular}{lrrrrrr}
\toprule
Model & Overall & 1--4 & 5--8 & 9--12 & 13--16 & 17--20 \\
\midrule
MP-20 train (GT) & 100.00 & 100.00 & 100.00 & 100.00 & 100.00 & 100.00 \\
\midrule
\textbf{GLASS (1M)} & \textbf{99.96} & \textbf{100.00} & 99.96 & \textbf{100.00} & \textbf{99.90} & \textbf{99.92} \\
\textbf{GLASS (50k)} & 92.07 & 99.28 & 94.28 & 91.24 & 85.12 & 88.72 \\
Crystalite & 97.51 & 99.95 & 99.70 & 98.77 & 95.50 & 91.90 \\
ADiT & 91.29 & 99.80 & 97.75 & 91.68 & 77.23 & 84.52 \\
Zatom-1 & 74.55 & 99.60 & 96.25 & 77.73 & 49.57 & 34.61 \\
MatterGen & 99.91 & 99.95 & 99.89 & 99.97 & 99.79 & 99.78 \\
OMatG & 94.65 & 99.90 & 98.83 & 93.33 & 91.16 & 87.39 \\
DiffCSP & 99.02 & 99.95 & \textbf{100.00} & 99.63 & 99.00 & 95.81 \\
\bottomrule
\end{tabular}%
}
\end{minipage}
\vspace{0.8em}

\begin{minipage}[t]{0.49\textwidth}
\centering
\textbf{Charge validity}\\[2pt]
\resizebox{\linewidth}{!}{%
\begin{tabular}{lrrrrrr}
\toprule
Model & Overall & 1--4 & 5--8 & 9--12 & 13--16 & 17--20 \\
\midrule
MP-20 train (GT) & 98.45 & 98.83 & 97.83 & 98.63 & 98.18 & 98.90 \\
\midrule
\textbf{GLASS (1M)} & 98.42 & 98.83 & 97.68 & 98.55 & 98.35 & 98.81 \\
\textbf{GLASS (50k)} & 96.62 & 98.08 & 95.36 & 96.66 & 95.79 & 97.49 \\
Crystalite & 96.85 & 98.17 & 96.26 & 95.97 & 96.90 & 97.40 \\
ADiT & \textbf{98.81} & 99.14 & 97.62 & \textbf{98.81} & \textbf{99.38} & \textbf{99.62} \\
Zatom-1 & 98.56 & \textbf{99.35} & \textbf{97.86} & 98.61 & 98.34 & 98.74 \\
MatterGen & 96.13 & 97.96 & 95.71 & 95.57 & 96.09 & 95.88 \\
OMatG & 96.13 & 98.29 & 96.45 & 93.91 & 95.79 & 96.62 \\
DiffCSP & 96.05 & 98.17 & 96.61 & 94.69 & 95.99 & 94.93 \\
\bottomrule
\end{tabular}%
}
\end{minipage}
\hfill
\begin{minipage}[t]{0.49\textwidth}
\centering
\textbf{Plausibility validity}\\[2pt]
\resizebox{\linewidth}{!}{%
\begin{tabular}{lrrrrrr}
\toprule
Model & Overall & 1--4 & 5--8 & 9--12 & 13--16 & 17--20 \\
\midrule
MP-20 train (GT) & 100.00 & 100.00 & 100.00 & 100.00 & 100.00 & 100.00 \\
\midrule
\textbf{GLASS (1M)} & \textbf{100.00} & \textbf{100.00} & \textbf{100.00} & \textbf{100.00} & \textbf{100.00} & \textbf{100.00} \\
\textbf{GLASS (50k)} & 99.99 & 99.97 & \textbf{100.00} & \textbf{100.00} & \textbf{100.00} & \textbf{100.00} \\
Crystalite & \textbf{100.00} & \textbf{100.00} & \textbf{100.00} & \textbf{100.00} & \textbf{100.00} & \textbf{100.00} \\
ADiT & \textbf{100.00} & \textbf{100.00} & \textbf{100.00} & \textbf{100.00} & \textbf{100.00} & \textbf{100.00} \\
Zatom-1 & \textbf{100.00} & \textbf{100.00} & \textbf{100.00} & \textbf{100.00} & \textbf{100.00} & \textbf{100.00} \\
MatterGen & \textbf{100.00} & \textbf{100.00} & \textbf{100.00} & \textbf{100.00} & \textbf{100.00} & \textbf{100.00} \\
OMatG & 99.93 & 99.66 & \textbf{100.00} & \textbf{100.00} & \textbf{100.00} & \textbf{100.00} \\
DiffCSP & 99.97 & \textbf{100.00} & \textbf{100.00} & 99.96 & \textbf{100.00} & 99.88 \\
\bottomrule
\end{tabular}%
}
\end{minipage}
\end{table}

\FloatBarrier
\Needspace{0.5\textheight}
\subsection{Crystalite-Protocol Results}
\label{app:crystalite_evaluation_results}

Table~\ref{tab:crystalite-dng} uses the separate protocol in Appendix~\ref{sec:crystalite_evaluation}. GLASS closely matches the density distribution but overproduces distinct elements: $3.70$ per structure on average versus $3.01$ in MP20 training data, explaining its larger $W_{\mathrm{N\!\text{-}ary}}$.

\begin{table}[H]
\centering
\caption{\textbf{Crystalite-protocol evaluation on MP20.} GLASS uses $10,000$ samples per run across three training runs at 50k. Sampling times are reported for 1,000 structures.}
\label{tab:crystalite-dng}
\small
\setlength{\tabcolsep}{3.2pt}
\glassfit{%
\begin{tabular}{lrrrrrrrrrr}
\toprule
Model & Struct. val. & Comp. val. & Unique & Novel & U.N. & Stable & S.U.N. & $W_\rho$ & $W_{\mathrm{N\!\text{-}ary}}$ & Time/1k \\
& (\%) $\uparrow$ & (\%) $\uparrow$ & (\%) $\uparrow$ & (\%) $\uparrow$ & (\%) $\uparrow$ & (\%) $\uparrow$ & (\%) $\uparrow$ & $\downarrow$ & $\downarrow$ & (s) $\downarrow$ \\
\midrule
MP20 train & 100.00 & 90.41 & -- & -- & -- & -- & -- & -- & -- & -- \\
\midrule
FlowMM & 93.03 & 83.15 & 97.44 & 85.00 & 83.99 & 46.05 & 31.64 & 1.389 & \textbf{0.075} & 1560 \\
CrystalDiT & 77.82 & 67.28 & 90.88 & 59.33 & 56.86 & \textbf{83.41} & 41.70 & 0.202 & 0.171 & 73.72 \\
DiffCSP & \textbf{99.93} & 82.10 & 96.90 & 89.53 & 87.89 & 50.28 & 38.60 & 0.192 & 0.344 & 237 \\
MatterGen & 99.78 & 83.72 & \textbf{98.10} & \textbf{91.14} & \textbf{90.26} & 51.70 & 42.29 & 0.088 & 0.184 & 2639 \\
ADiT & 99.52 & \textbf{90.15} & 90.25 & 59.80 & 56.91 & 76.90 & 36.76 & 0.231 & 0.089 & 84.81 \\
\midrule
Crystalite & 99.61 & 81.72 & 95.24 & 79.30 & 77.33 & 69.72 & \textbf{47.49} & \textbf{0.051} & 0.127 & 22.36 / 5.14$^\dagger$ \\
& $\pm$0.06 & $\pm$0.24 & $\pm$0.19 & $\pm$0.12 & $\pm$0.21 & $\pm$0.85 & $\pm$0.77 & $\pm$0.010 & $\pm$0.006 &  \\
\midrule
GLASS (ours) & 99.57 & 86.02 & 96.36 & 72.45 & 71.47 & 67.79 & 39.58 & 0.0667 & 0.703 & 0.21 \\
 & $\pm$ 0.08 & $\pm$ 0.05 & $\pm$ 0.06 & $\pm$ 0.68 & $\pm$ 0.78 & $\pm$ 1.07 & $\pm$ 0.30 & $\pm$ 0.0060 & $\pm$ 0.033 &  \\
\bottomrule
\end{tabular}
}

{\footnotesize $^\dagger$ Crystalite reports 22.36\,s for the original and 5.14\,s for the optimized implementation. GLASS timing uses one full H100 (Table~\ref{tab:sampling_cost}).}
\end{table}

\FloatBarrier
\subsection{Crystalite Sampling Ablation}
\label{app:crystalite-churn}

We sample the released Crystalite MP20 model with and without noise re-injection (churn) and coordinate/lattice anti-annealing. Each setting uses $10,000$ samples, EMA weights, 150 Heun steps on the Karras schedule \citep{karrasElucidatingDesignSpace2022}, empirical MP20 atom counts, $S_{\mathrm{churn}}=60$ when enabled, and $S_{\mathrm{noise}}=1.003$.

With anti-annealing disabled, enabling churn raises LeMat validity from $67.03\,\%$ to $94.39\,\%$, and from $16.67\,\%$ to $89.48\,\%$ at 17--20 atoms. Anti-annealing changes overall validity by at most $0.21$ percentage points.

\begin{table}[H]
\centering
\caption{\textbf{Crystalite sampler settings on MP20.} LeMat-GenBench total validity (\%) for the four Crystalite sampler settings on 10,000 matched MP-20 samples per setting. Churn uses $S_{\mathrm{churn}}=60$; anti-annealing is applied to coordinates and lattices with $\rho=10$. Bold values are best in each column.}
\label{tab:crystalite-sampler-lemat-validity}
\small
\begin{tabular}{lccrrrrrr}
\toprule
Sampler & Churn & Anti-ann. & Overall & 1--4 & 5--8 & 9--12 & 13--16 & 17--20 \\
\midrule
Vanilla & No & No & 67.03 & \textbf{98.17} & 92.65 & 73.25 & 37.29 & 16.67 \\
Anti-annealing only & No & Yes & 66.82 & 98.11 & 92.26 & 73.13 & 37.59 & 15.90 \\
Churn only & Yes & No & 94.39 & \textbf{98.17} & 95.79 & \textbf{94.86} & \textbf{92.40} & 89.48 \\
Production & Yes & Yes & \textbf{94.44} & 98.11 & \textbf{96.00} & \textbf{94.86} & \textbf{92.40} & \textbf{89.54} \\
\bottomrule
\end{tabular}
\end{table}

\FloatBarrier
\Needspace{0.75\textheight}
\section{Additional QMOF150 Results}
\label{sec:qmof_additional}

\FloatBarrier
\subsection{Structural Validity and Relaxation}

GLASS and QMOF150 training structures use the official MOFChecker 0.9.6. Overall validity requires carbon, hydrogen, and a metal, with none of the defect flags in Table~\ref{tab:qmof-mofchecker-full}. Defect frequencies can overlap. Three-dimensional graph connectivity is not required for this validity definition. Failed evaluations count as invalid.

GLASS results use $10,000$ samples from each of three training runs. The size curves pool these samples and show 95\% Wilson intervals. Mofasa curves use the precomputed MOFChecker flags for $10,000$ row-matched raw and relaxed MofasaDB samples, selected uniformly among raw structures with 20--150 atoms. Mofasa uses a different training split with structures up to 170 atoms. The table retains published baseline results, including Orb-v3+D3 relaxation for Mofasa-opt.

eSEN-OAM+D3 relaxation gives a mean displacement RMSD of $0.2028 \pm 0.0029$\,\AA{} over $30,000$ structures across three training runs.

\begin{table}[H]
\centering
\caption{\textbf{Full MOFChecker breakdown (\%).} Presence and overall validity are pass rates; other rows are defect frequencies. Opt denotes relaxed structures. $^*$Mofasa uses a different training split and size range; table values retain published evaluations. Bold values indicate best-in-class, separate for raw and optimized structures.}
\label{tab:qmof-mofchecker-full}
\begingroup
\small
\setlength{\tabcolsep}{4pt}
\renewcommand{\arraystretch}{1}
\glassfit{%
  \begin{tabular}{lrrrrrrrrr}
    \toprule
    Criterion & Ref. & \multicolumn{2}{c}{ADiT} & Zatom-1 & SinAE & \multicolumn{2}{c}{Mofasa$^*$} & \multicolumn{2}{c}{GLASS (1M)} \\
    \cmidrule(lr){2-2} \cmidrule(lr){3-4} \cmidrule(lr){5-5} \cmidrule(lr){6-6} \cmidrule(lr){7-8} \cmidrule(lr){9-10}
    & Train & QMOF & Joint & & & Raw & Opt & Raw & Opt \\
    \midrule
Has carbon $\uparrow$ & 100.0 & \textbf{100.0} & \textbf{100.0} & \textbf{100.0} & \textbf{100.0} & 98.4 & \textbf{100.0} & 99.1 & \textbf{100.0} \\
Has hydrogen $\uparrow$ & 99.8 & 99.6 & \textbf{100.0} & \textbf{100.0} & 99.8 & 98.4 & \textbf{99.9} & 98.9 & 99.8 \\
Atomic overlap $\downarrow$ & 0.0 & 8.3 & 10.8 & 8.8 & 5.2 & 2.8 & \textbf{0.0} & \textbf{1.3} & 0.1 \\
Overcoordinated C $\downarrow$ & 0.0 & 23.6 & 34.3 & 1.1 & 12.8 & 2.5 & \textbf{0.0} & \textbf{0.5} & \textbf{0.0} \\
Overcoordinated N $\downarrow$ & 0.0 & 1.5 & 1.6 & \textbf{0.0} & 0.4 & 1.6 & \textbf{0.0} & 0.1 & \textbf{0.0} \\
Overcoordinated H $\downarrow$ & 0.0 & \textbf{1.0} & 3.6 & 2.9 & 1.3 & 2.4 & \textbf{0.0} & \textbf{1.0} & \textbf{0.0} \\
Undercoordinated C $\downarrow$ & 5.9 & 60.0 & 72.1 & 65.4 & 39.2 & 20.1 & 14.3 & \textbf{8.2} & \textbf{8.2} \\
Undercoordinated N $\downarrow$ & 6.9 & 39.1 & 39.9 & 22.8 & 26.9 & 12.9 & \textbf{7.3} & \textbf{11.4} & 8.1 \\
Undercoordinated rare earth $\downarrow$ & 0.0 & 0.4 & 0.8 & \textbf{0.0} & 0.2 & 2.1 & 0.2 & 0.1 & \textbf{0.1} \\
Has metal $\uparrow$ & 100.0 & \textbf{100.0} & 99.4 & \textbf{100.0} & \textbf{100.0} & 97.7 & 99.2 & 99.0 & \textbf{99.8} \\
Lone molecule $\downarrow$ & 9.7 & 72.9 & 83.2 & 80.5 & 49.8 & 31.3 & 27.1 & \textbf{12.1} & \textbf{11.8} \\
High charge $\downarrow$ & 1.0 & 0.9 & 2.5 & 0.5 & \textbf{0.2} & 3.5 & 1.6 & 1.4 & \textbf{1.1} \\
Suspicious terminal oxo $\downarrow$ & 0.0 & 2.6 & 5.8 & \textbf{0.3} & 3.1 & 2.5 & 0.5 & 0.5 & \textbf{0.0} \\
Undercoordinated alkali/alkaline $\downarrow$ & 0.2 & 1.0 & 6.4 & \textbf{0.3} & 0.7 & 3.0 & 1.0 & \textbf{0.3} & \textbf{0.4} \\
Geometrically exposed metal $\downarrow$ & 1.7 & 7.0 & 9.6 & \textbf{1.8} & 4.0 & 5.3 & 3.7 & 2.6 & \textbf{3.0} \\
Overall valid $\uparrow$ & 80.3 & 15.7 & 10.2 & 15.1 & 16.3 & 52.9 & 59.8 & \textbf{74.8} & \textbf{78.7} \\
    \bottomrule
  \end{tabular}%
  }
\endgroup
\end{table}

\FloatBarrier
\Needspace{0.6\textheight}
\subsection{Validity--Novelty Tradeoff}
\label{sec:qmof_mofid_protocol}
\label{sec:qmof_checkpoint_sweep}

The permissive MOFid convention used by Mofasa retains unresolved topologies such as \texttt{ERROR}, \texttt{NA}, and \texttt{UNKNOWN}; the strict convention requires a complete topology-resolved identifier. They identify $84.89\,\%$ and $32.02\,\%$ of QMOF150 training structures, respectively. Novelty is the absence of an identifiable training reference under the same convention. Missing reference identifiers can make recalled structures appear novel.

All yields use the requested sample count. Unidentified candidates do not count as novel. Valid-and-novel yield counts candidates passing both criteria; VNU additionally removes duplicate identifiers, counting an identifier when at least one candidate is valid. Conditional validity among identified novel samples uses that subset as its denominator.

\begin{figure}[H]
\centering
\includegraphics[width=.49\linewidth]{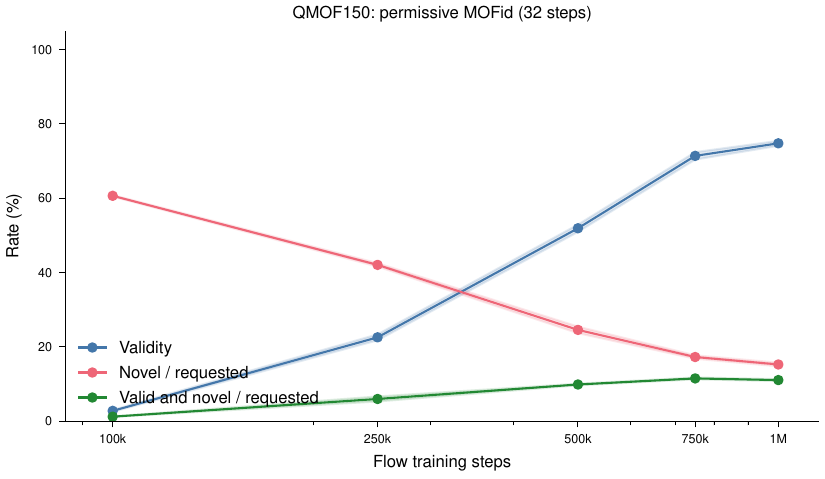}
\includegraphics[width=.49\linewidth]{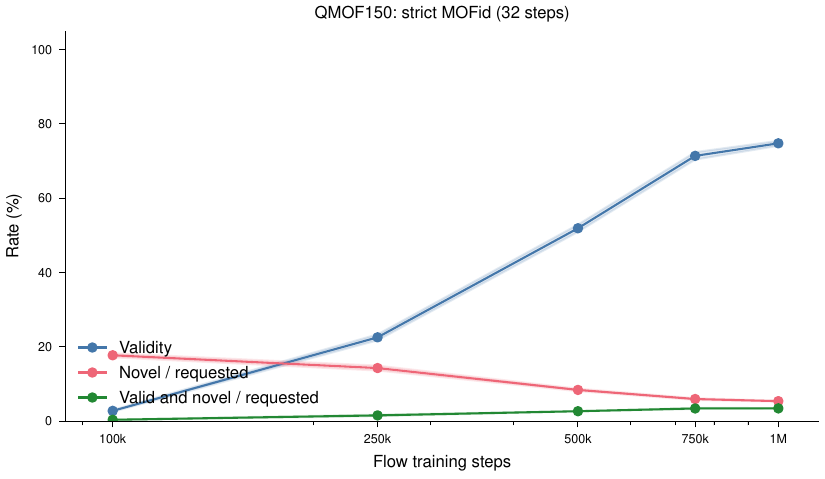}
\caption{\textbf{QMOF150 validity--novelty tradeoff.} Raw validity, MOFid novelty, and valid-and-novel yield under permissive (left) and strict (right) identifiers. Means and sample standard deviations over three training runs, with $10,000$ samples per run and checkpoint; all rates use requested-sample denominators.}
\label{fig:qmof-checkpoint-tradeoff}
\end{figure}

Validity rises and novelty falls during flow training (Figure~\ref{fig:qmof-checkpoint-tradeoff}). At 1M steps, permissive VNU is $\QmofVnu\,\%$ and strict VNU is $\QmofStrictVnu\,\%$. Mofasa reports $42.4\,\%$ under the same permissive convention, with a different training split and reference \citep{simkusMofasaStepChange2025}. Its higher reported novel yield accompanies lower raw structural validity.

\begin{figure}[H]
\centering
\includegraphics[width=.65\linewidth]{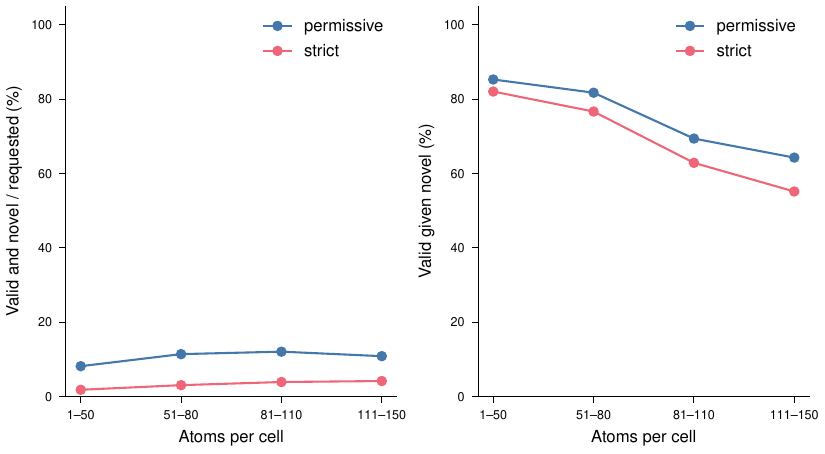}
\caption{\textbf{QMOF150 novel yield by size at 1M steps.} Valid-and-novel yield and validity among identified novel candidates under each MOFid convention. Identifier novelty can include close variants of training frameworks.}
\label{fig:qmof_novelty_size}
\end{figure}

Table~\ref{tab:qmof150-latent-mofid-protocols} extends the sweep to latent dimensions 32 and 128. The validity--novelty tradeoff persists across widths.

\begin{table}[H]
\centering
\caption{\textbf{QMOF150 generation by latent dimension and flow checkpoint.} One training run per setting, with $10,000$ raw samples per checkpoint. Rates use the requested-sample denominator.}
\label{tab:qmof150-latent-mofid-protocols}
\label{tab:qmof150-latent-ablation}
\label{tab:qmof150-latent-ablation-permissive}
\begingroup
\small
\setlength{\tabcolsep}{4pt}
\renewcommand{\arraystretch}{1}
\glassfit{%
\begin{tabular}{rrr rrrr rrrr}
\toprule
& & &
\multicolumn{4}{c}{Mofasa-permissive MOFid} &
\multicolumn{4}{c}{Strict complete MOFid} \\
\cmidrule(lr){4-7}\cmidrule(lr){8-11}
Latent dim. & Flow step & $V$ &
$E$ & $N$ & $U$ & $VNU$ &
$E$ & $N$ & $U$ & $VNU$ \\
\midrule
32 & 100k & 3.13 & 67.85 & 60.25 & 66.48 & 1.32 & 20.37 & 17.35 & 20.04 & 0.44 \\
64 & 100k & 2.88 & 68.60 & \textbf{60.45} & 67.15 & 1.29 & 21.56 & \textbf{18.25} & 21.18 & 0.33 \\
128 & 100k & 2.44 & 61.11 & 55.65 & 60.05 & 1.03 & 18.41 & 16.23 & 18.15 & 0.35 \\
32 & 250k & 18.54 & 80.54 & 47.76 & \textbf{72.13} & 5.14 & 28.58 & 15.62 & 25.76 & 1.32 \\
64 & 250k & 23.45 & 80.74 & 41.60 & 69.46 & 6.44 & 30.21 & 15.04 & \textbf{26.20} & 1.69 \\
128 & 250k & 16.90 & 77.18 & 47.84 & 70.24 & 4.33 & 26.82 & 15.35 & 24.53 & 1.33 \\
32 & 500k & 47.19 & 84.29 & 28.75 & 65.93 & 9.04 & 31.41 & 9.47 & 24.78 & 2.19 \\
64 & 500k & 52.99 & 82.29 & 23.57 & 62.85 & 9.11 & 31.15 & 8.07 & 24.25 & 2.50 \\
128 & 500k & 44.82 & 81.76 & 29.62 & 65.43 & 8.06 & 29.70 & 9.89 & 23.91 & 2.33 \\
32 & 750k & 70.72 & 85.01 & 18.48 & 60.75 & \textbf{9.67} & \textbf{32.26} & 6.34 & 23.57 & 2.85 \\
64 & 750k & 72.56 & 83.33 & 16.88 & 60.06 & 9.45 & 31.47 & 6.06 & 23.13 & 2.87 \\
128 & 750k & 64.26 & 83.75 & 20.42 & 62.26 & 8.95 & 30.87 & 7.21 & 23.32 & 2.77 \\
32 & 1M & 73.14 & \textbf{85.20} & 16.84 & 59.88 & 9.07 & 32.07 & 5.63 & 23.12 & 2.68 \\
64 & 1M & \textbf{75.54} & 83.19 & 14.73 & 59.44 & 8.62 & 31.47 & 5.34 & 22.96 & 2.69 \\
128 & 1M & 69.67 & 84.19 & 17.89 & 61.33 & 9.14 & 31.21 & 6.38 & 23.27 & \textbf{2.97} \\
\bottomrule
\end{tabular}%
}
\endgroup
\medskip
\begin{minipage}{0.98\linewidth}
\footnotesize
$E$ is the fraction with an available identifier; $V$ is the MOFChecker-valid fraction; $N$ is the fraction whose identifier is absent from the corresponding QMOF150 training-reference set; $U$ is the number of distinct generated identifiers divided by the requested sample count; and $VNU$ is the number of distinct identifiers that are both valid and novel, divided by the requested sample count.
\end{minipage}
\end{table}

\FloatBarrier
\subsection{Similarity to Training Structures}
\label{sec:pipeline_localization}

Before flow fitting, QMOF150 reconstruction RMSD is \QmofTrainCellRmsd\,\AA\ on training structures and \QmofValCellRmsd\,\AA\ on held-out structures. Increasing latent dimension does not close this gap (Appendix~\ref{sec:qmof_capacity}). During flow training, the decoder is fixed, while generated samples increasingly match training frameworks (Table~\ref{tab:qmof-training-matches}).

\begin{figure}[H]
\centering
\includegraphics[width=\linewidth]{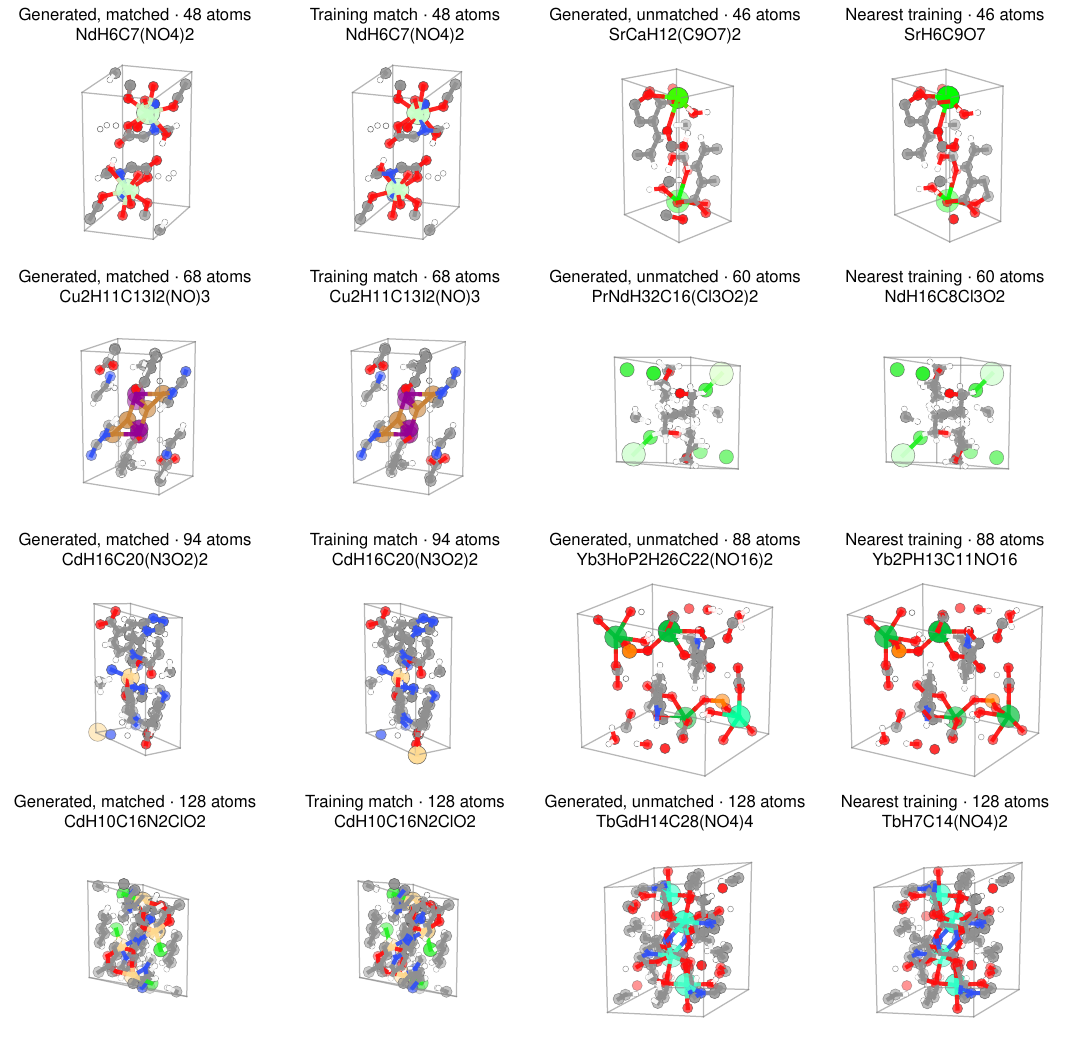}
\caption{\textbf{Generated QMOF150 structures and training analogues.} Each row covers one size range. Left: a valid generation and its StructureMatcher training match. Right: a valid unmatched generation and its nearest fingerprint analogue. These unmatched examples retain the training framework with substituted metal sites.}
\label{fig:qmof_generated_analogues}
\end{figure}

We compare valid generations with all training structures of the same reduced formula using pymatgen's StructureMatcher \citep{ongPythonMaterialsGenomics2013}: element identity, site tolerance 0.3, lattice tolerance 0.2, angle tolerance $5^\circ$, primitive-cell reduction, cell scaling, and supercell matching. At 1M steps, \QmofUnmatchedNewFormula\,\% of unmatched valid GLASS samples have a reduced formula absent from training. The unmatched examples in Figure~\ref{fig:qmof_generated_analogues} retain training frameworks with metal substitutions.

For Mofasa, we also search QMOF150 validation and test structures because its training split differs. No additional matches are found: \MofasaAnySplitMatchRate\,\% match any of the $16,637$ QMOF150 structures. This comparison uses the existing local MOFChecker 0.9.6 compatibility-profile scores for its valid subset; the size curves instead use MofasaDB's stored scores.

\begin{table}[H]
\centering
\caption{\textbf{StructureMatcher comparison of valid generations with QMOF150 training structures.} GLASS values are means over three training runs with $10,000$ raw samples each. Validity and valid-unmatched yield use all requested samples; the match rate is conditional on validity. $^*$10,000 released samples; different training split and size range.}
\label{tab:qmof-training-matches}
\small
\begin{tabular}{lrrr}
\toprule
Flow step & Valid (\%) & Matched $\mid$ valid (\%) & Valid, unmatched (\%) \\
\midrule
250k & 22.57 & 90.27 & 2.20 \\
500k & 51.93 & 97.00 & 1.56 \\
1M & \textbf{74.79} & 98.29 & 1.28 \\
Mofasa$^*$ & 54.03 & \textbf{0.15} & \textbf{53.95} \\
\bottomrule
\end{tabular}
\end{table}

\FloatBarrier
\Needspace{0.55\textheight}
\section{Latent Representations and Slot Organization}
\label{sec:latent_analysis}

The PCA projections below show qualitative organization of the training latents; axis labels give explained variance.

\begin{figure}[H]
    \centering
    \includegraphics[width=\linewidth]{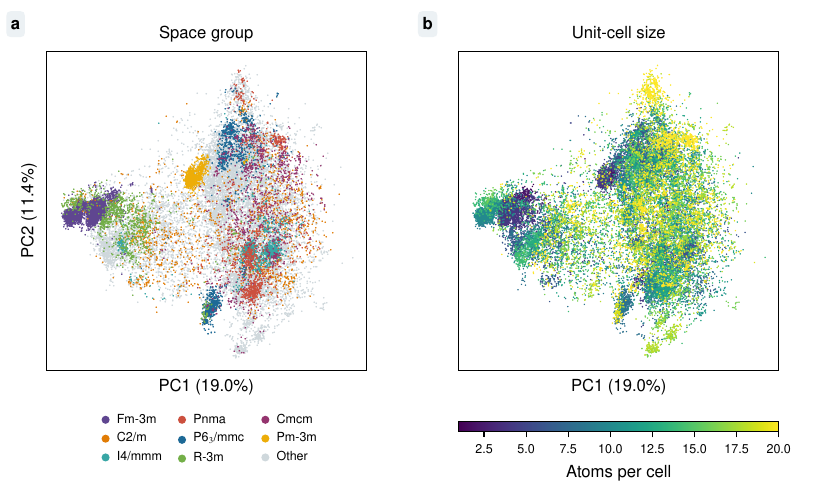}
    \caption{\textbf{MP20 latent-space visualization.} PCA of MP20 autoencoder training latents, colored by space group (a) and atom count (b). Axis labels report explained variance.}
    \label{fig:mp20_latent_pca}
\end{figure}

\begin{figure}[H]
    \centering
    \includegraphics[width=\linewidth]{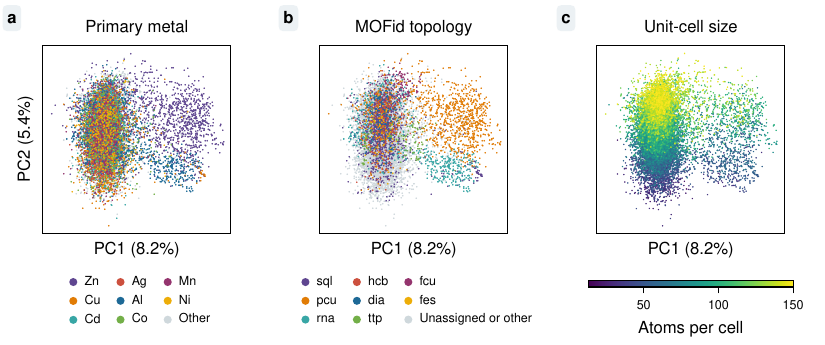}
    \caption{\textbf{QMOF150 training latents.} PCA colored by the most frequent metal (a), permissive MOFid net (b), and atom count (c); grey denotes unassigned nets. One region contains predominantly Zn/\texttt{pcu} and Al/\texttt{rna} frameworks, while other classes overlap.}
    \label{fig:qmof_latent_pca}
\end{figure}

\label{sec:slot_correspondence}
\begin{figure}[H]
    \centering
    \includegraphics[width=\linewidth]{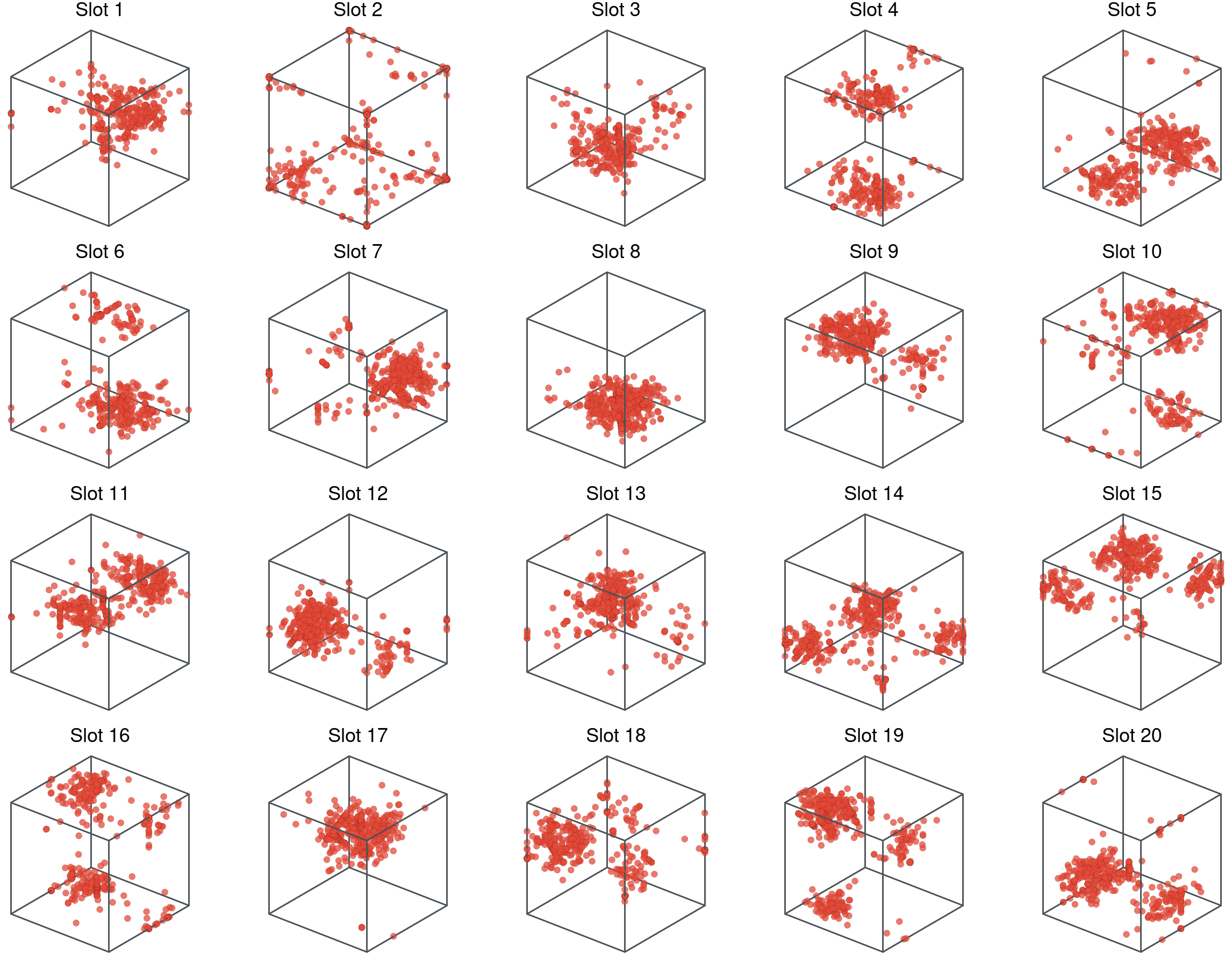}
    \caption{\textbf{Learned MP20 slot correspondences.} Atoms assigned to each decoder slot across $1,000$ training structures, showing the fractional coordinates. Empty slots are excluded. Spatial patterns indicate specialization within the supplied coordinate frame.}
    \label{fig:all_slots_1000_grid}
\end{figure}

\FloatBarrier
\Needspace{0.8\textheight}
\section{Computational Cost}
\label{sec:computational_cost}

For dense-attention particle models, $K_x$ model evaluations cost $K_x C_{\mathrm{atom}}(n)$, where $C_{\mathrm{atom}}(n)=\mathcal O(n^2)$ at fixed width and depth. GLASS costs
\begin{equation}
T_{\mathrm{GLASS}}=K_z C_{\mathrm{latent}}(d_z)+C_{\mathrm{dec}}(M).
\end{equation}
Latent integration is independent of atom count at fixed capacity; the $\mathcal O(M^2)$ slot decoder runs once. Hungarian matching is used during autoencoder training only.

On one NVIDIA H100, autoencoder training takes about 20 minutes for MP20 and 3.5 hours for QMOF150; 1M flow steps take about 50 minutes on either dataset.

Table~\ref{tab:sampling_cost} reports medians over five repetitions after batch-size-specific warm-up on one NVIDIA H100 GPU. A 32-step midpoint trajectory uses 64 flow evaluations and one decoder pass. Total time includes device-to-host transfer; flow and decoder timings use GPU events. Loading, serialization, relaxation, and evaluation are excluded. The main text reports the lowest median across the listed batch sizes.

\begin{table}[H]
\centering
\caption{\textbf{Sampling cost per 1,000 structures on one full H100.} Median seconds over five repetitions; peak memory is allocated GPU memory.}
\label{tab:sampling_cost}
\small
\begin{tabular}{lrrrrr}
\toprule
Dataset & Batch size & Total (s) & Flow (s) & Decoder (s) & Peak memory (GiB) \\
\midrule
MP20 & 32 & 1.74 & 1.72 & 0.022 & 0.43 \\
 & 128 & 0.67 & 0.66 & 0.012 & 0.46 \\
 & 256 & 0.39 & 0.38 & 0.009 & 0.50 \\
 & 512 & 0.27 & 0.26 & 0.009 & 0.57 \\
 & 1,000 & 0.21 & 0.20 & 0.008 & 0.71 \\
\midrule
QMOF150 & 32 & 1.86 & 1.71 & 0.139 & 0.53 \\
 & 128 & 0.78 & 0.65 & 0.127 & 0.74 \\
 & 256 & 0.51 & 0.38 & 0.123 & 1.02 \\
 & 512 & 0.38 & 0.26 & 0.117 & 1.55 \\
 & 1,000 & 0.32 & 0.20 & 0.118 & 2.60 \\
\bottomrule
\end{tabular}
\end{table}

\clearpage
\FloatBarrier
\section{Generated Structures}
\label{app:generated-structures}

Examples are selected from valid raw generations across atom-count ranges and shown with training analogues. StructureMatcher identifies matches; unmatched candidates are compared by normalized composition and minimum-image pair-distance fingerprints over the complete training split.

\begin{figure}[H]
\centering
\includegraphics[width=\linewidth]{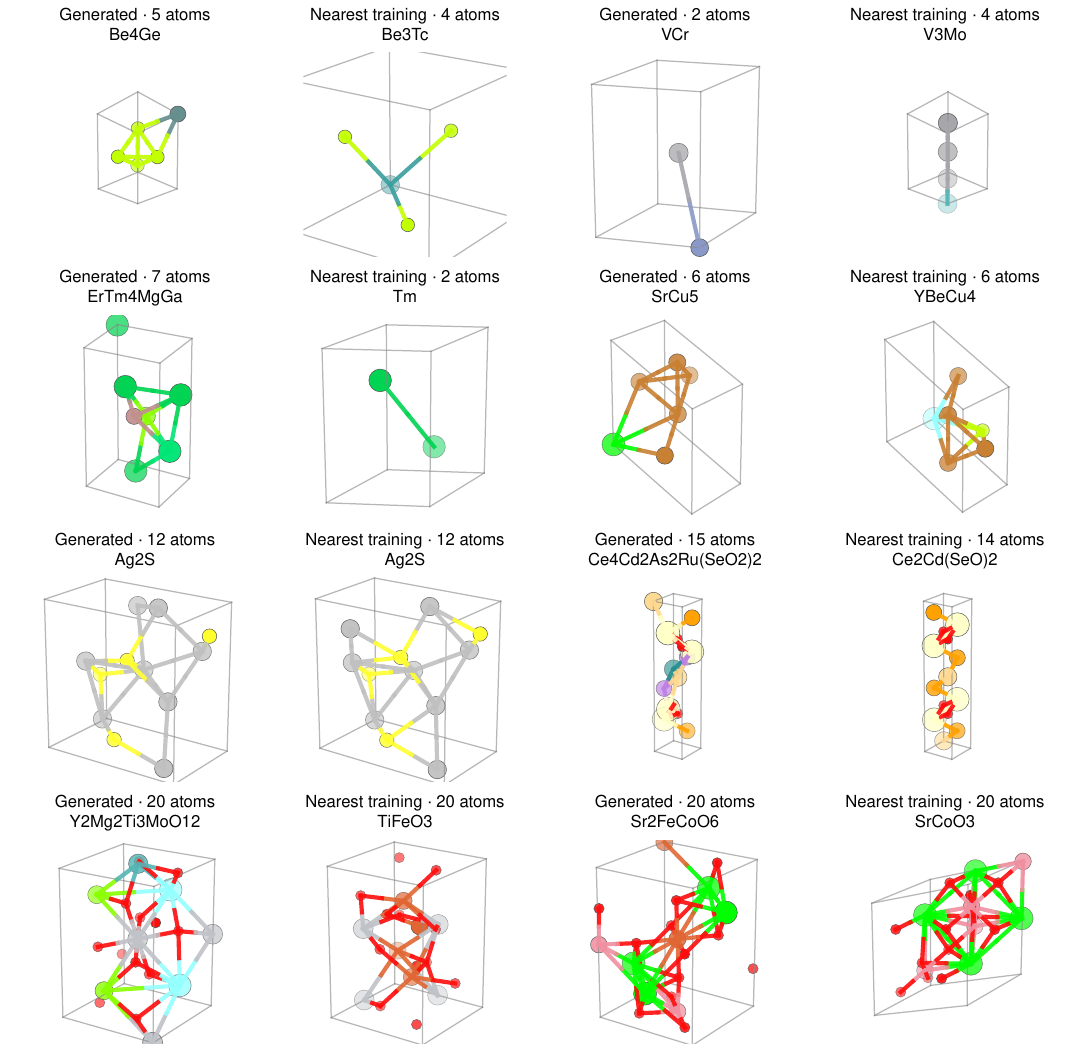}
\caption{\textbf{Generated MP20 structures and training analogues.} Two valid examples per size bin (1--5, 6--10, 11--15, and 16--20 atoms), each beside its nearest training structure under the composition and pair-distance fingerprint. Bonds are distance-based visual guides.}
\label{fig:generated_analogues}
\end{figure}

\end{document}